%% file: main.tex
\documentclass[10pt,twocolumn,letterpaper]{article}

\usepackage[accsupp]{axessibility} 
\usepackage{iccv}
\usepackage{times}
\usepackage{epsfig}
\usepackage{graphicx}
\usepackage{amsmath}
\usepackage{amssymb}

\usepackage{color, colortbl,xcolor}
\definecolor{Gray}{gray}{0.9}
\definecolor{light-gray}{gray}{0.95}

\usepackage{pifont}%
\newcommand{\cmark}{\ding{51}}%
\newcommand{\xmark}{\ding{55}}%

\usepackage{booktabs} %
\newcommand{\ra}[1]{\renewcommand{\arraystretch}{#1}} %
\usepackage{multirow} %
\usepackage{subcaption} %
\usepackage{url} %

\usepackage{microtype}
\usepackage{cuted} %
\usepackage{capt-of}
\usepackage[font={small}]{caption}
\usepackage{epigraph}
\usepackage{comment}
\usepackage{dsfont}
\usepackage{enumitem}%

\NewDocumentCommand{\tjd}{ mO{} }{\textcolor{red}{\textsuperscript{\textit{Trevor}}\textsf{\textbf{\small[#1]}}}}
\NewDocumentCommand{\sg}{ mO{} }{\textcolor{magenta}{\textsuperscript{\textit{Shiry}}\textsf{\textbf{\small[#1]}}}}
\usepackage[pagebackref=true,breaklinks=true,letterpaper=true,colorlinks,bookmarks=false]{hyperref}

\usepackage[capitalize]{cleveref}
\crefname{section}{Sec.}{Secs.}
\Crefname{section}{Section}{Sections}
\Crefname{table}{Table}{Tables}
\crefname{table}{Tab.}{Tabs.}

\iccvfinalcopy %

\def\iccvPaperID{3663} %

\ificcvfinal\fi

\begin{document}

\title{Can Language Models Learn to Listen?}

\author{Evonne Ng\thanks{denotes equal contribution}\quad
Sanjay Subramanian\footnotemark[1] \quad
Dan Klein\quad
Angjoo Kanazawa\quad
Trevor Darrell\quad
Shiry Ginosar \\
\\
University of California, Berkeley
}

\maketitle
\ificcvfinal\thispagestyle{empty}\fi

\begin{strip}\centering
\includegraphics[width=\linewidth]{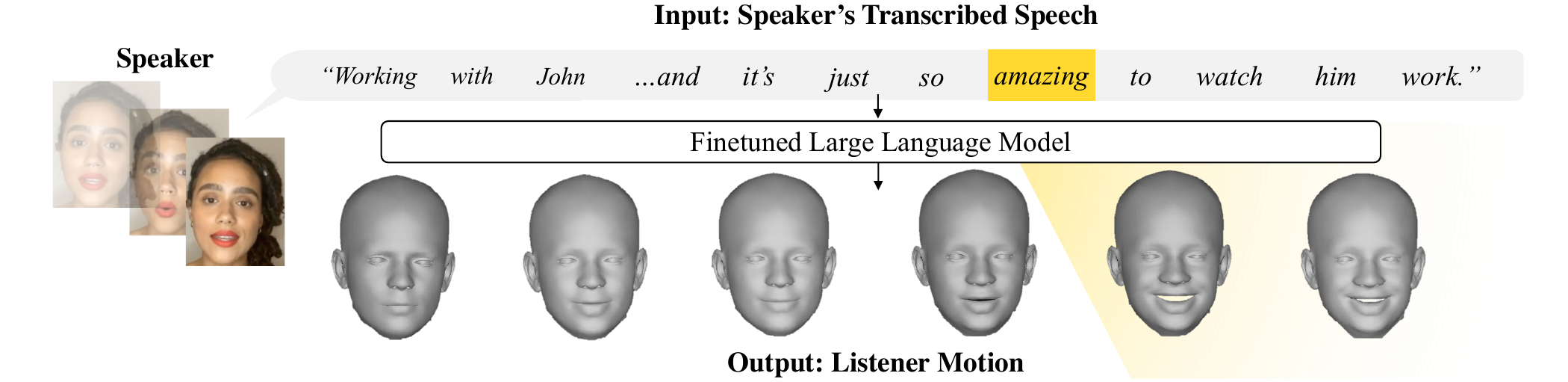}
\captionof{figure}{{\bf Large language models transfer to listener motion prediction.}
Given a video of a listener and speaker pair, we extract text corresponding to the spoken words of the speaker. We fine-tune a pretrained large language model to autoregressively generate realistic 3D listener motion in response to the input transcript. Our method generates semantically meaningful gestures (\eg~an appropriately timed smile inferred from ``amazing") that synchronously flow with the conversation.
We can optionally render the output of our approach as photorealistic video.
Video: \small{\url{https://youtu.be/djpSOhdIU8M}}
}
\label{fig:teaser}
\end{strip}

\begin{abstract}
We present a framework for generating appropriate facial responses from a listener in dyadic social interactions based on the speaker's words. Given an input transcription of the speaker's words with their timestamps, our approach autoregressively predicts a response of a listener: a sequence of listener facial gestures, quantized using a VQ-VAE. 
Since gesture is a language component, we propose treating the quantized atomic motion elements as additional language token inputs to a transformer-based large language model. Initializing our transformer with the weights of a language model pre-trained only on text results in significantly higher quality listener responses than training a transformer from scratch. We show that our generated listener motion is fluent and reflective of language semantics through quantitative metrics and a qualitative user study. In our evaluation, we analyze the model's ability to utilize temporal and semantic aspects of spoken text.
\end{abstract}

\input{sections/intro}

\input{sections/related_work}

\input{sections/methods}

\input{sections/results}

\medskip {\footnotesize
\noindent \textbf{Acknowledgements.} The work of Ng, Subramanian and Darrell is supported by BAIR's industrial alliance programs, and the DoD DARPA's Machine Common Sense and/or SemaFor programs. Subramanian is also supported by an NDSEG fellowship. Ginosar's work is funded by NSF under Grant \# 2030859 to the Computing Research Association for CIFellows Project.}
{\small
\bibliographystyle{ieee_fullname}
\bibliography{main}
}

\clearpage

{\Large{\textbf{Appendix}}}
\appendix

\input{supp}

\end{document}

%% file: sections/intro.tex
\section{Introduction}

\begin{figure*}[t]
    \centering
    \includegraphics[width=\textwidth]{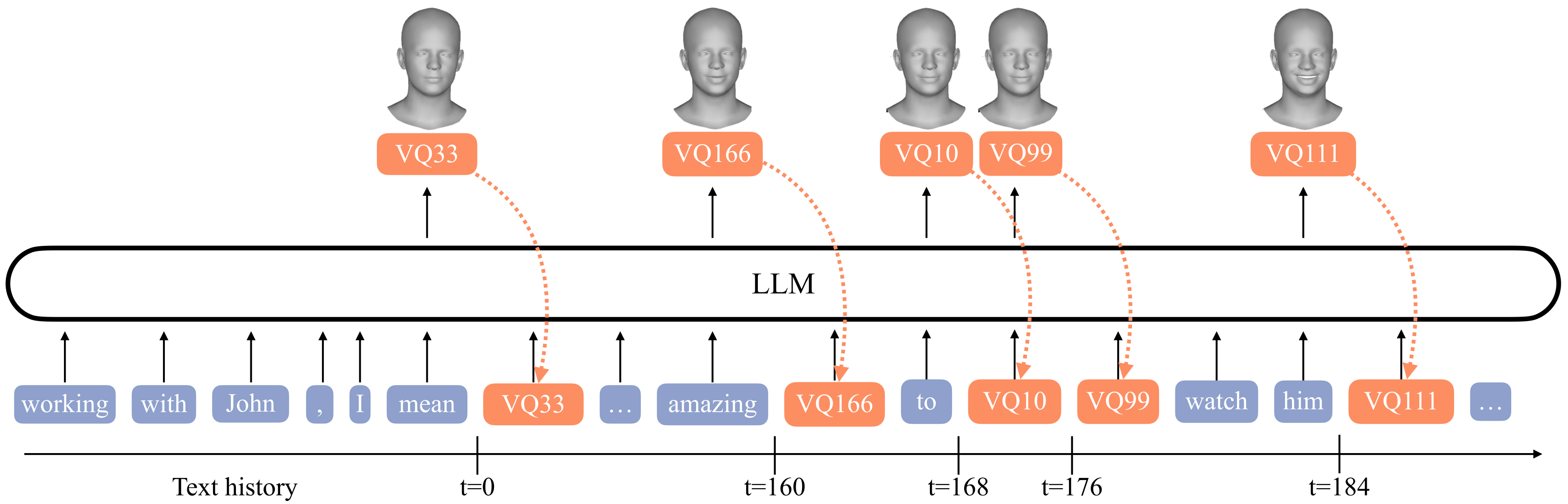}
    \caption{\textbf{Listener motion prediction model.} The model takes as input text tokens (blue), along with their timestamps, and predicts tokens representing atomic listener motion elements (orange) that we discretized with a VQ-VAE. We feed in a fixed-size history window of text tokens before the listener response's onset. Then we generate one discrete gesture token at a time while providing text tokens as the speaker speaks (\ie according to word timestamps). $t$ denotes the number of frames that have elapsed since the start of motion generation. Each discrete motion token represents $8$ frames of continuous motion.}
    \label{fig:predictor}
\end{figure*}

Human face-to-face communication is multifaceted and multimodal~\cite{condon_ogston_1966,kendon1970movement}. In particular, the flow and effectiveness of face-to-face dyadic interactions critically depend on non-verbal motions and responses from both participants in the conversation~\cite{kendon1970movement,LaFrance1979,chartrand1999chameleon}. This paper focuses on the non-verbal facial feedback listeners provide to speakers during a dyadic conversation~\cite{ng2022learning,zhou2022responsive,geng2023affective}.

When listening to a speaker, we produce gestures in response to several multimodal communication channels, including speech, non-verbal gestures, and lexical semantics (the meaning of words)~\cite{kendon1970movement}.  
Previous studies have shown that speech and gesture are informative of listener responses~\cite{ng2022learning,zhou2022responsive}. Here we ask how far we can get with lexical semantics alone. This question is significant given the abundant availability of textual dialogue data, in contrast to the limited availability of conversational motion datasets. To study the transferability of large language models to the dyadic conversational motion domain, we propose the task depicted in Figure~\ref{fig:teaser} of predicting listener motion from the raw text transcribed from the speaker's words.

Since gesture is a conversational communication channel, our central insight is to transfer knowledge from pretrained large language models to the gesture generation task. We propose to treat discrete atomic motion elements as novel language tokens.
We first learn a data-driven dictionary of discrete atomic gesture elements by training a VQ-VAE~\cite{vqvae} to capture the full spectrum of videotaped listener responses~\cite{ng2022learning}. We then fine-tune a pretrained large language model to autoregressively predict these novel motion tokens given temporally-aligned speaker text (see Figure~\ref{fig:predictor}). We ensure that each motion token is only generated based on previously spoken words by interleaving the input speaker text tokens with the autoregressively predicted motion tokens. Hence, our causal model can produce listener responses in real-time as it does not rely on future speaker words.

Our text-conditioned model outperforms baselines in both quantitative metrics and human evaluation. Notably, it performs competitively with prior work that uses audio and visual input for listener generation. The generated listener responses are consistently on-par with ground truth listener facial gestures, as our perceptual study demonstrates.

Given these results, we ask why a text-conditioned model performs well on an inherently multimodal task. We focus our analysis of the model on the two main qualities we expect from realistic listener non-verbal feedback: (1) temporally synchronous responses, such as head nods, and (2) semantic, emotionally meaningful responses, such as smiles or looks of puzzlement~\cite{CassellThorisson1999}. We find that a text transcription of a speaker's utterance carries some temporal signal of when a response is in order. Punctuation, capitalization, and temporal breaks in word delivery are hints about sentence structure that embed this rhythmic information.
We further demonstrate that lexical semantics is crucial for producing the correct emotional response, especially when the speaker's facial expression does not reflect the emotional affect of their words. Finally, we note that, as expected, our model cannot capture responses for which the facial affect or motion of the speaker is crucial.

%% file: sections/related_work.tex
\section{Related Work}

\medskip \noindent \textbf{Conversational dynamics}
Works on animating conversational avatars traditionally involved crafting rule-based designs on lab-captured motion data~\cite{Cassell1994, gratch2006virtual, huang2011virtual, Bohus_Horvitz_2010, sonlu2021conversational}, which often limit the variety of captured gestures, or rely on simplifying assumptions for motion generation that do not hold for in-the-wild data. As a result, several data-driven approaches were proposed for tasks such as predicting the head pose of the speaker and listener~\cite{greenwood2017predicting}, turn taking~\cite{joo2019towards, ahuja2019react}, and single-frame facial expressions that summarize the sequence~\cite{huang2017dyadgan,nojavanasghari2018interactive}. While these methods focus on a narrow aspect of modeling social dynamics, our approach captures the natural complexity of interactions by considering the full range of facial expressions and head rotations.

More recently, there have been works on modeling a speaker's fine-grain motion generation conditioned on audio~\cite{jonell2019learning, ginosar2019learning},~\cite{chu2018face} text, or both audio and text~\cite{kucherenko2020gesticulator}. However, all these works focus on modeling monadic conversational settings where the goal is to output speaker motion that directly matches the input signals. In contrast,~\cite{learn2smile2017, jonell2019learning, ng2022learning, zhou2022responsive} model cross-person, dyadic interactions by predicting the listener's 2D~\cite{learn2smile2017} or 3D motion~\cite{jonell2019learning,ng2022learning,zhou2022responsive}. Yet all these prior works condition the listener's response on the speaker's motion and audio. In addition,~\cite{zhou2022responsive} relies on a one-bit semantic affect conditioning that signifies whether the synthesized listener should have a positive, negative, or neutral response. In contrast, we focus on demonstrating that semantically meaningful and temporally realistic responses that correspond to a given speaker are possible from a text-only context. 

Most related to our work is~\cite{geng2023affective}, which explicitly models the semantics within a conversation. By prompting GPT3~\cite{brown2020language} with a predefined goal and text of the speaker,~\cite{geng2023affective} obtains visual details of the listener from which they train a model to retrieve listener clips that most closely match this description within their dataset. Rather than taking the full input text at once, our model ingests time-aligned text and autoregressively outputs corresponding 3D listener motion. Additionally, we \emph{generate} raw 3D motion rather than picking from existing clips, which allows us to derive motion that does not exist in the training set.
All these differences allow us to model \emph{both} semantics and temporally realistic responses in dyadic conversations.

\medskip \noindent \textbf{Text driven motion synthesis} 
Several prior works considered text-conditioned 3D motion generation~\cite{petrovich2022temos, tevet2022motionclip, tevet2022human, guo2022generating,t2mgpt}. \cite{tevet2022motionclip, tevet2022human} leverage existing pretrained large language models to produce semantically meaningful embeddings as input to their system. In contrast to these methods, we consider text-conditioned motion in conversational settings. We explore the potential for using pretrained large language models to discover semantic and temporal information from a speaker's transcript. We demonstrate that from text alone, we can generate temporally aligned motion indicative of synchronous responses in conversational settings. 

Recent work showed that knowledge from large language models can transfer to other tasks by finetuning pretrained models~\cite{lu2021pretrained, reid2022can, liu2023language}. We leverage this insight and demonstrate that finetuning on a pretrained large language model transfers well to conversational motion generation.

%% file: sections/methods.tex
\section{Listener Motion Generation with LLMs}
Given the speaker's transcribed speech in a dyadic conversation, we aim to generate corresponding listener facial motions. Our system consists of two components: (1) a model that converts listener motion into a sequence of discrete tokens and (2) an autoregressive model that predicts future motion tokens conditioned on previously generated motion tokens and previously spoken words.

\subsection{Problem Definition}
Let $\mathbf{F} = \{\mathbf{f}_1, \mathbf{f}_2, ..., \mathbf{f}_T\}$ represent the listener's face across $T$ frames during one of the speaker's turns in the conversation. Let $W = (w_1, w_2, ..., w_N)$ be a sequence of text tokens corresponding to the words spoken during the time spanned by frames $1$ to $T$. Let $M = (m_1, m_2, ..., m_N)$ be the corresponding timestamps, where $m_i \in \{1, 2, ..., T\}$ denotes the timestamp of the frame corresponding to the end of the interval in which token $w_i$ was spoken. For each pair of frames $t_1 < t_2$, we denote by $W_{t_1:t_2}$ the sequence of words spoken between frames $t_1$ and $t_2$. 

Additionally, we consider some historical text context corresponding to words the speaker said earlier in their turn, before frame $1$. Let $W^{history} = (w^{history}_1, w^{history}_2, ..., w^{history}_{N'})$ be the sequence of text tokens spoken during the $H$ seconds of the speaker's turn before frame $1$.

Our generator $\mathcal{G}$ takes as input $W$, $W^{history}$, and $M$ and predicts $\mathbf{F}$. Specifically, the $t$-th predicted face in the sequence is given by
\begin{align*}
    \hat{\mathbf{f}}_t =&\; \mathcal{G}(W^{history}, W_{1:t}, M_{1:t}, \mathbf{F}_{1:t-1}).
\end{align*}

To estimate 3D facial expressions and orientations from video frames of human conversations, we represent the face of the listener in every frame using a 3D Morphable Face Model (3DMM)~\cite{blanz1999morphable,paysan20093d,cao2013facewarehouse, FLAME:SiggraphAsia2017}. 3DMMs are parametric facial models that allow us to directly regress disentangled coefficients corresponding to facial expression, head orientation, and identity-specific shape from a single image~\cite{zollhofer2018state}. We obtain facial expression coefficients $\beta_t \in \mathds{R}^{d_m}$, where $d_m$ is the dimension of the expression coefficient, a normalized 3D head pose $R_t \in SO(3)$, and shape coefficients that we discard to obtain an identity-agnostic representation. 
Our facial representation at time $t$, $\mathbf{f}_t \in \mathds{R}^{d_m+3}$, is a concatenation of expression and orientation (in Euler angles):
\begin{equation}
    \mathbf{f}_t = [\beta_t,  R_t].
\end{equation}

\subsection{Discretizing Listener Motion}
Predicting the sequence of facial expressions $\mathbf{F}$ is challenging because the coefficients of $\mathbf{f}_t$ are real-valued and require regression-based methods. Following recent work in motion generation~\cite{ng2022learning,t2mgpt}, we use a VQ-VAE~\cite{vqvae} to encode a sequence of facial expressions into a sequence of discrete tokens. We can then predict these discrete tokens via straightforward classification-based methods.

The VQ-VAE consists of an \emph{encoder} neural network, a \emph{decoder} neural network, and a set of \emph{codebook} embeddings $\mathbf{C} \in \mathbb{R}^{V_{vq} \times d_c}$, where $V_{vq}$ is the size of the codebook and $d_c$ is the dimension of the embeddings. Each codebook embedding corresponds to a unique discrete token in the codebook. The encoder takes as input the sequence of facial expressions $F^L = (\mathbf{f}_1, \mathbf{f}_2, ..., \mathbf{f}_T)$, normalized by the mean and standard deviation across all training examples, and produces as output a sequence of latent features $\mathbf{Z} = (\mathbf{z}_1, \mathbf{z}_2, ..., \mathbf{z}_{T/r})$, where $r$ is the downsampling rate of the encoder and each $\mathbf{z}_i \in \mathbb{R}^{d_c}$. The \emph{quantizer} $Q$ is a deterministic, parameter-free function that converts each vector of this sequence to a codebook token:
\begin{align*}
    Q(\mathbf{z}_i) =&\; \arg \min_{1 \leq j \leq V} ||z_i-\mathbf{C}_j||^2
\end{align*}
We denote by $q_i = Q(\mathbf{z}_i)$ the $i$th VQ token.
The decoder takes as input a sequence of codebook embeddings
$(\mathbf{C}_1, \mathbf{C}_2, \ldots, \mathbf{C}_{T/r})$
and, after reverse normalization by mean and standard deviation, produces as output a continuous sequence of facial expressions
$\hat{F}^L = (\hat{\mathbf{f}}_1, \hat{\mathbf{f}}_2, \ldots, \hat{\mathbf{f}}_3)$.

\medskip \noindent \textbf{Architecture and Training}
The architectures of the VQ-VAE encoder and decoder mainly consist of convolutional and residual layers and are shown in %
Appendix Figure~\ref{fig:vqvae}.
We train the VQ-VAE with a combination of four losses:
\begin{align*}
    L_{embed} =&\; \sum_{t=1}^{T/r} ||\mathbf{z}_t-sg[C_{q_t}]||^2 \\
    L_{reconstruct} =&\; \sum_{t=1}^{T} \mathcal{L}_1^{smooth}(\hat{\mathbf{f}}_t, \mathbf{f}_t) \\
    L_{velocity} =&\; \sum_{t=1}^{T-1} \mathcal{L}_1^{smooth}(\hat{\mathbf{f}}_{t+1}-\hat{\mathbf{f}}_t, \mathbf{f}_{t+1}-\mathbf{f}_{t})
\end{align*}
Here $sg$ denotes the stop-gradient operator, and $\mathcal{L}_1^{smooth}$ denotes the L1 smooth loss function. The total training loss is a weighted sum of these four losses. We also use exponential moving average and codebook reset when training.

\subsection{Text-conditioned Motion Generation}
Our autoregressive motion generation model $\mathcal{G}$ outputs listener facial responses in two steps. First, we predict a series of discrete codebook tokens. Second, we decode these discrete tokens via the VQ-VAE decoder to a sequence of 3DMM coefficient vectors representing continuous motion. Figure~\ref{fig:predictor} illustrates our architecture.

We instantiate $\mathcal{G}$ with a language model (LM) based on a transformer architecture~\cite{transformer}. Specifically, we use GPT2~\cite{gpt2}. A LM takes as input a sequence of text tokens $\{w_1, w_2, ..., w_H\}$ and outputs a distribution over the vocabulary for the next token. The first layer in the LM is an embedding $E_{word} \in \mathbb{R}^{V_{word} \times d_w}$, where $V_w$ is the vocabulary size and $d_w$ is the embedding dimension, that converts the token indices to embeddings $\{\mathbf{e}_1, \mathbf{e}_2, ..., \mathbf{e}_H\}$. We use a positional embedding matrix $P \times \mathbb{R}^{H \times d_w}$ to add positional information to each token's embedding: $\mathbf{e}'_i = \mathbf{e}_i + P_i$.

Each of the remaining layers in the model takes a sequence of vectors $\{\mathbf{a}_1, \mathbf{a}_2, ..., \mathbf{a}_H\}$ which can be represented as a matrix $A \in \mathbb{R}^{H \times d_w}$ and uses linear projections on $A$ to produce \emph{query}, \emph{key}, and \emph{value} matrices $Q, K, V \in \mathbb{R}^{H \times d_w}$. Let $L_T$ be a $H \times H$ matrix in which elements on and below the diagonal are $1$, and all other elements are $-\infty$. Then for each position in the sequence, self-attention is computed between $Q$ and $K$ to compute a distribution over the sequence, and the output representation $A'$ is a weighted sum of the value vectors according to this distribution:
\begin{align*}
    \alpha =&\; \textrm{softmax}\left(L_T \odot \frac{QK^T}{\sqrt{d_w}}\right) \\
    A' =&\; \alpha V
\end{align*}
Here $\odot$ denotes the element-wise product. In contrast to a bidirectional transformer where each position attends to all others, this \emph{causal} attention mechanism ensures that each position attends only to itself and previous ones. Finally, we apply LayerNorm and a feedforward network to $A'$ to produce the input to the following layer. The final layer of the LM is an affine projection that predicts the next token in the sequence.

\begin{table*}\centering \footnotesize
\setlength{\tabcolsep}{12.0pt}.
\ra{1.3} %
\begin{tabular}{@{}lrrrrrr@{}}
\toprule %
& L2 $\downarrow$ & FD $\downarrow$ & variation & diversity & P-FD $\downarrow$ & L2 Affect ($10^2$) $\downarrow$ \\
\cmidrule{2-7}
\rowcolor{light-gray}
\textit{GT} & & & $0.11$ &  $2.59$ & & \\
Random Train & $0.63 \pm 0.02$ & $30.35 \pm 1.0$ & $0.088 \pm 0.005$ & $2.26 \pm 0.06$ & $31.47 \pm 1.0$ & $11.91 \pm 0.73$ \\
Random VQ & $0.71 \pm 0.01$ & $29.31 \pm 0.6$ & $0.269 \pm 0.004$ & $4.83 \pm 0.05$ & $31.44 \pm 0.6$ & $10.14 \pm 0.53$ \\
NN & $0.52 \pm 0.02$ & $23.68 \pm 1.1$ & $0.087 \pm 0.004$ & $2.25 \pm 0.05$ & $24.78 \pm 1.1$ & $7.88 \pm 0.56$ \\
Uncond & $\mathbf{0.39} \pm 0.02$ & $21.28 \pm 0.9$ & $0.002 \pm 0.000$ & $0.42 \pm 0.00$ & $21.65 \pm 0.9$ & $7.63 \pm 0.56$ \\
\rowcolor{Gray}
Full & $0.43 \pm 0.02$ & $\mathbf{18.22} \pm 0.7$ & $0.116 \pm 0.005$ & $2.81 \pm 0.06$ & $\mathbf{19.63} \pm 0.8$ & $\mathbf{6.36} \pm 0.47$ \\
\bottomrule
\end{tabular}
\caption{\textbf{Results.} Comparison against ground truth annotations (GT). $\downarrow$ indicates lower is better; closer to GT is better for no arrow. We average each metric over the test set instances. Standard error is computed via bootstrap (using 10,000 samples).}
\label{tab:results}
\end{table*}

To enable the LM to generate motion VQ tokens, we instantiate randomly initialized word embeddings $E_{VQ} \in \mathbb{R}^{V_{vq} \times d_w}$ for each of the VQ tokens, and we replace the output layer with a randomly initialized affine projection that outputs logits for each of the VQ tokens. 
We determine the order of the text and VQ tokens by ensuring that for each VQ token, the previous text tokens are those whose timestamp is less than the timestamp of the current VQ token. 

Often a listener's reaction is not only determined by what is immediately being said but also by what has already been said. To model this additional context in the conversation, we also include history tokens for the text that has occurred before the first frame at $t=1$. The first $N'$ tokens in the input are the text history tokens $W^{history}$. The listener motion tokens $q_1, q_2, ..., q_{T/r}$ are placed in order after the history tokens. The text tokens spoken during the segment are placed as follows: for each $t \in \{1, 2, ..., T/r-1\}$, the set of text tokens that are placed between $q_t$ and $q_{t+1}$ is $W_{rt:r(t+1)}$. We also place a space token (\ie, the textual space token from the GPT2 tokenizer) just before each listener motion token.

As in text-only LMs, we train the model using cross-entropy loss on the task of next-token prediction with teacher-forcing:
\begin{align*}
    \mathcal{L} =&\; -\sum_{t=1}^{T/r} \log\Pr\left[\mathcal{G}(W^{history}, W_{1:rt}, q_{1:t-1}) = q_t\right]
\end{align*}
At test time, we use greedy decoding to predict the sequence of motion tokens.

%% file: sections/results.tex
\section{Experimental Setup}
We conduct quantitative experiments to validate the effectiveness of our method in comparison with baselines (section~\ref{sec:quant-results}). We then demonstrate, via an A/B test on Mechanical Turk, that our results are comparable to ground truth and preferable to those produced by baselines (section~\ref{sec:mturk}).

We then analyze why the text-conditioned model performs well on this multimodal task. We examine the two qualities expected from listener non-verbal feedback: (1) temporally synchronous (sec.~\ref{sec:temporal-analysis}) and (2) semantically-meaningful (sec.~\ref{sec:semantic-analysis}) responses.
We demonstrate that text carries some temporal information that informs when a response is appropriate. We further find that lexical semantics is crucial for producing the correct emotional response. Finally, we show that more historical text context improves performance, but too much context leads to degradation~\ref{sec:history-analysis}. For implementation details, see 
Appendix~\ref{app:method}.

\subsection{Dataset}
As in prior work on listener motion generation \cite{ng2022learning}, we focus on the person-specific modeling setting, in which all videos share the same listener. We evaluate on one of the listeners (Trevor Noah) in the dataset introduced by~\cite{ng2022learning}. We improve this dataset in three ways. First, we segment the raw videos into longer segments.  The original dataset included 2-second segments, but more context is needed here since text is a sparse signal.  We train and evaluate on segments of up to 8 seconds (not including the length of the text history provided to the model). To identify the listener segments, we use PyAnnote to perform speech diarization \cite{pyannote1,pyannote2}. Second, to extract 3DMMs from the videos, we use EMOCA~\cite{emoca,emocav2}, a more expressive model than DECA~\cite{deca}, used for the original dataset. Finally, we extract time-aligned speech transcriptions using Whisper~\cite{whisper}. During training and testing, we only include motion that starts at least 3 seconds after the start of the speaker's turn, and only consider segments that are at least $24$ frames. This procedure results in $2366$ training, $222$ validation, and $543$ test segments.

\begin{table*}\centering \footnotesize
\setlength{\tabcolsep}{3.5pt}.
\ra{1.3} 
\begin{tabular}{@{}lrrrrcrrrrrrr@{}}
\toprule 
& PT GPT & align & ordered & text type && L2 $\downarrow$ & FD $\downarrow$ & variation & diversity & P-FD $\downarrow$ & L2 Affect ($10^2$) $\downarrow$ \\
\cmidrule{2-5} \cmidrule{7-12}
\rowcolor{light-gray}
\textit{GT} & & & & & && & $0.11$ &  $2.59$ & & \\
NoPT  & \xmark & \cmark & \cmark & given && $0.53 \pm 0.02$ & $22.81 \pm 0.9$ & $0.14 \pm 0.005$ & $3.21 \pm 0.06$ & $24.38 \pm 0.9$ & $8.41 \pm 0.53$ \\
Unaligned  & \cmark & \xmark & \cmark & given && $0.45 \pm 0.02$ & $19.03 \pm 0.8$ & $0.11 \pm 0.005$ & $2.71 \pm 0.06$ & $20.40 \pm 0.8$ & $6.66 \pm 0.51$  \\
Scrambled  & \cmark & \cmark & \xmark & given && $0.49 \pm 0.02$ & $19.96 \pm 0.7$ & $0.12 \pm 0.004$ & $2.86 \pm 0.06$ & $21.59 \pm 0.8$ & $7.37 \pm 0.50$ \\
FixTok  & \cmark & \cmark & \cmark & fixed && $0.73 \pm 0.02$ & $37.31 \pm 1.2$ & $0.05 \pm 0.003$ & $1.58 \pm 0.04$ & $38.36 \pm 1.2$ & $13.45 \pm 0.86$ \\
FixTok-Punc  & \cmark & \cmark & \cmark & punc.~+fixed && $0.60 \pm 0.02$ & $29.07 \pm 1.1$ & $0.09 \pm 0.005$ & $2.31 \pm 0.06$ & $30.41 \pm 1.1$ & $10.18 \pm 0.65$ \\
\rowcolor{Gray}
Full & \cmark & \cmark & \cmark & given && $\mathbf{0.43} \pm 0.02$ & $\mathbf{18.22} \pm 0.7$ & $0.12 \pm 0.005$ & $2.81 \pm 0.06$ & $\mathbf{19.63} \pm 0.8$ & $\mathbf{6.36} \pm 0.47$ \\
\bottomrule
\end{tabular}
\caption{\textbf{Ablations.} Each metric is averaged over the test set instances. Standard error is computed via bootstrap (using 10,000 samples).}
\label{tab:ablation}
\end{table*}

\subsection{Evaluation Metrics} Due to the difficulty of quantitatively evaluating realism in multimodal motion generation, we use an extensive suite of metrics that evaluate our predictions along multiple axes. Inspired by prior work~\cite{ng2022learning}, we focus on assessing our predicted listeners' realism, diversity, and synchrony with speaker motion.

\begin{itemize}
    \item \emph{L2}: on ground truth expression coefficients and pose
    \item \emph{Frechet distance (FD) for realism}: Motion realism measured by distribution distance between generated and ground-truth motion. We calculate FD~\cite{heusel2017gans} in the expression $\mathds{R}^{T \times d_m}$ and head pose  $\mathds{R}^{T \times 3}$ space of the full motion sequence.
    \item \emph{Variation for diversity}: Variance calculated across the sequence of expression coefficients or 3D rotations.
    \item \emph{Diversity}: Following~\cite{zhang2023generating}, we randomly sample 30 pairs of listener pose and expression parameters within a sequence of motion and compute the average Euclidean distances between the pairs to measure motion diversity in the set.
    \item \emph{Paired FD for synchrony}: 
    Quality of listener-speaker dynamics measured by distribution distances on listener-speaker \emph{pairs} (P-FD). FD~\cite{heusel2017gans} on concatenated listener-speaker motion $\mathds{R}^{T \times (d_m+d_m)}$/ pose $\mathds{R}^{T \times (3+3)}$.  
    \item \emph{L2 Affect for synchrony}: Measures the accuracy of the produced listener facial affect across the sequence. We average listener facial affect over a 1-second window and compute the L2 against ground truth in a sliding window manner.
\end{itemize}


Together, these metrics measure both the semantic appropriateness and the temporal synchrony of gestures between a speaker and listener in conversation.

\medskip \noindent \textbf{Baselines} We compare to the following baselines:

\begin{itemize}
    \item \textbf{NN text}: A segment-search method commonly used for synthesis. Given input speaker text, we find its nearest neighbor from the training set and use its corresponding listener segment as the prediction. We use the \texttt{all-mpnet-base-v2} model from SentenceTransformers~\cite{reimers-2019-sentence-bert} to encode text, commonly used for text retrieval.\footnote{\url{https://www.sbert.net/index.html}} On the validation set, this model performs slightly better than GPT2-medium on L2/FD and slightly worse on L2 Affect.
    \item \textbf{Random}: Return a randomly-chosen sequence of a listener from the training set.
    \item \textbf{Uncond}: Unconditional model that learns to produce motion sequences without text conditioning. Note that since we use greedy decoding, this method produces the same motion sequence for a given sequence length.
    \item \textbf{NoPT}: Our method without GPT pretrained weights.
\end{itemize}

\section{Results}
Through quantitative experiments, we demonstrate that our proposed method outperforms all baselines. In a Mechanical Turk A/B test, we further show that our predictions realistically correspond to the speaker and are competitive with an existing approach for listener motion synthesis conditioned on the speaker's speech and gesture.

\subsection{Quantitative Results}
\label{sec:quant-results}
Table~\ref{tab:results} shows our proposed method outperforms all other baselines and that finetuning on GPT2 is crucial. Overall, \textbf{Full} achieves the best balance of performance across all the various metrics. According to the FD, our method produces motion that matches the distribution of the ground truth dataset. Furthermore, the motion produced is similar in variation and diversity to real motion. Most notably, we generate synchronous motion, as shown by our method's strong performance in P-FD and L2 Affect over all baselines. As a result, this suggests our model generates accurate facial expressions that match the dynamics of the conversation. 

Furthermore, we calculate the Shannon Index~\cite{ng2022learning} on facial gestures to measure the overall entropy of generated expressions. Ours (2.52) is similar to the ground truth (2.68), which shows a fair amount of diversity and demonstrates that our predictions do not simply collapse into two modes (smile/not smile).

While \textbf{Uncond} has a lower L2, it performs significantly worse in all other metrics. This confirms that our model successfully leverages the text input. Similarly, \textbf{NoPT} performs poorly across the board, suggesting that GPT2 finetuning is advantageous for our task. 

\begin{figure*}[t]
\begin{center}
    \includegraphics[width=\textwidth]{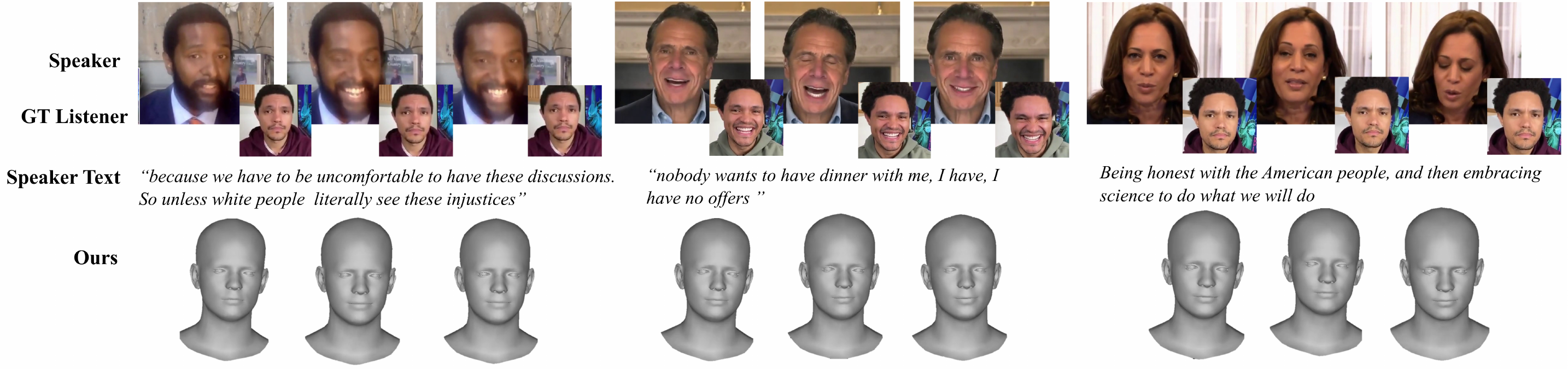}
    \caption{\textbf{Where we win and where we lose.} Our model responds in an emotionally-appropriate manner when lexical semantics is crucial. For example, when it is not appropriate to smile despite a speaker's uneasy laughter (Left). We fail to capture cases that can only rely on the speaker's facial motion, such as sarcastic jokes (Middle). In many cases, both the textual sentence structure and the speaker's gesture contain hints of \textit{when} a response, such as a nod, is appropriate, allowing us to model speaker-listener synchrony despite no access to motion (Right). See \small{\url{https://youtu.be/djpSOhdIU8M}} for a video version of these examples.}
    \label{fig:win-lose}
\end{center}
\end{figure*}





\subsection{Human Evaluation}
\label{sec:mturk}
To corroborate our quantitative results and gain insight into how our synthesized listeners perceptually compare to real motion
, we conducted an A/B test on Amazon Mechanical Turk. 
We visualized listener motion using videos of grayscale 3D facial meshes. 

Participants watched a series of video pairs. In each pair, our model generated one video; an ablation or a baseline produced the other.
Participants were then asked to identify the video containing the listener that looks like they are listening and paying \emph{more} attention to the speaker.
Videos of at least 8 seconds each of resolution $849 \times 450$ (downsampled from $1132 \times 600$ in order to fit two videos vertically stacked on different screen sizes) were shown, and after each pair, participants were given unlimited time to respond. 
Since the most tell-tale moments for when a listener is truly listening are during defining moments (speaker tells a joke, shares a sad story, \etc) that illicit strong responses, we manually curated $47$ such notable moment sequences from our held-out test data. We then predicted a corresponding listener 3D facial motion sequence using each method. For every test sequence, each A/B comparison was made by $3$ evaluators.

We compared our strongest baselines \textbf{NN} and \textbf{Uncond} to our proposed model and recorded the percentage of times participants preferred our method over the baseline models or vice versa. \textbf{Ours} significantly outperformed. $70.1\%$ of the total 150 evaluators preferred \textbf{Ours} over \textbf{NN}, and $92.8\%$ preferred \textbf{Ours} over \textbf{Uncond}. These statistics reflect the quantitative trends in Table~\ref{tab:results}.~Furthermore, in a comparison against avatars rendered from ground truth listeners, evaluators preferred \textbf{Ours} $49.7\%$ of the time. This \textit{highlights the perceptual realism of our predicted listener motion}.

Additionally, we compare against prior SOTA Learning to Listen(L2L)~\cite{ng2022learning} that models the temporal synchrony of a speaker-listener dyad from speech audio and motion. AMT evaluators preferred ours over L2L $55.7\%$ of the time. This suggests that our text-only approach models synchronous motion comparable to that of an approach that explicitly models temporal synchrony through prosody, which is known to encode the beats and pacing of a conversation through inflections, tones, and rhythm of speech. That said, we note that for this comparison we used the original L2L implementation, which relied on DECA~\cite{deca}, an older and less expressive 3DMM than EMOCA~\cite{emoca}.

\section{Analysis of the Text-based Method}
Responding to a speaker nonverbally is inherently a multimodal task~\cite{kendon1970movement}. And yet, we demonstrated that we could design a competitive method for this task that relies on text input alone. We now analyze our proposed approach to unearth some reasons for its success. We discuss the two qualities of non-verbal listening feedback corresponding to successful listening feedback~\cite{Cassell1994}: temporal synchrony with the speaker and semantically-appropriate responses. We then consider the effect of varying lengths of historical temporal context in the input text. We conclude this section by discussing the limitations of taking only text as input.

\begin{figure}[t]
\begin{center}
\includegraphics[width=\linewidth]{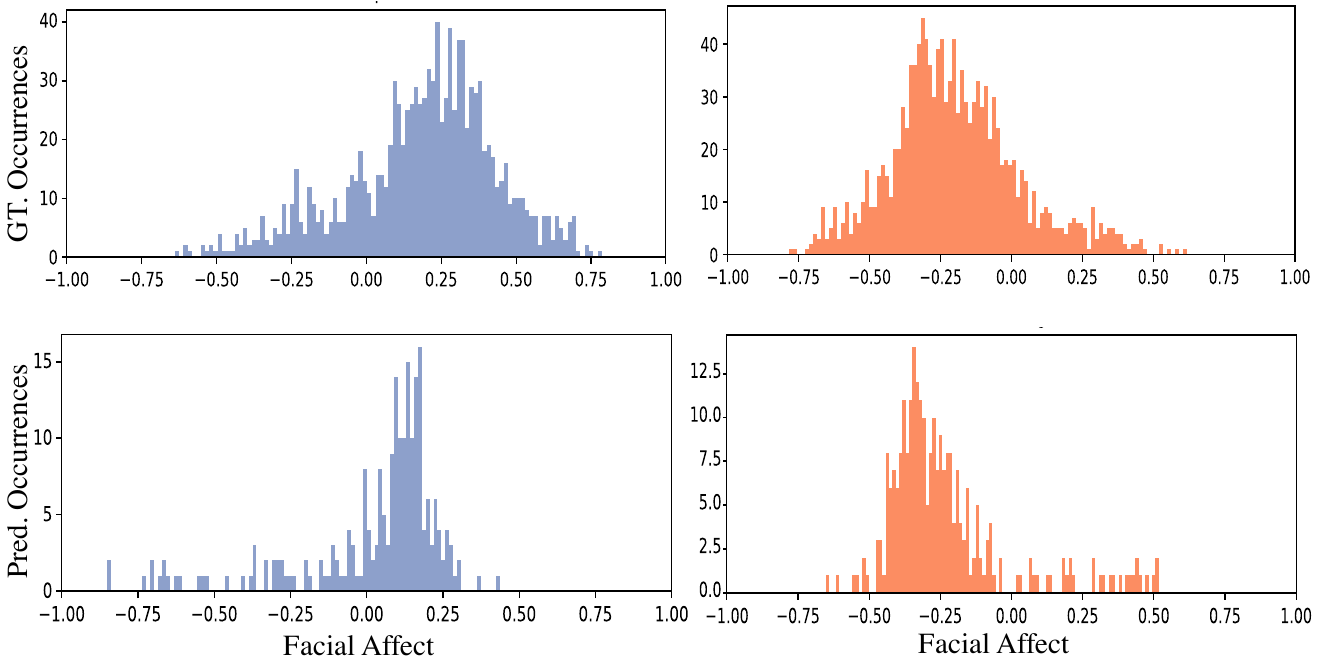}
\caption{\textbf{Positive phrases elicit positive affect, and vice versa.} Given the top 100 most positive (left) and negative (right) phrases, we plot a histogram of the facial affect of a listener during and 2 seconds after the stated phrase. -1 corresponds to very upset. The ground truth distribution (top), computed over all the data, and our predicted distribution (bottom), computed over the test data, exhibit a robust correlation.}
\label{fig:lsaffect}
\end{center}
\end{figure}

\begin{figure}[t]
\begin{center}
\includegraphics[width=0.80\linewidth]{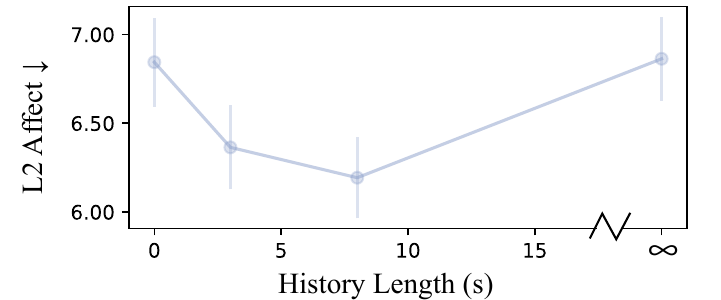}
\caption{\textbf{Effect of text history.} Providing text history helps, but too much history hurts the model in terms of affect.}
\label{fig:history}
\end{center}
\end{figure}

\begin{figure}[t]
\begin{center}
    \includegraphics[width=\linewidth]{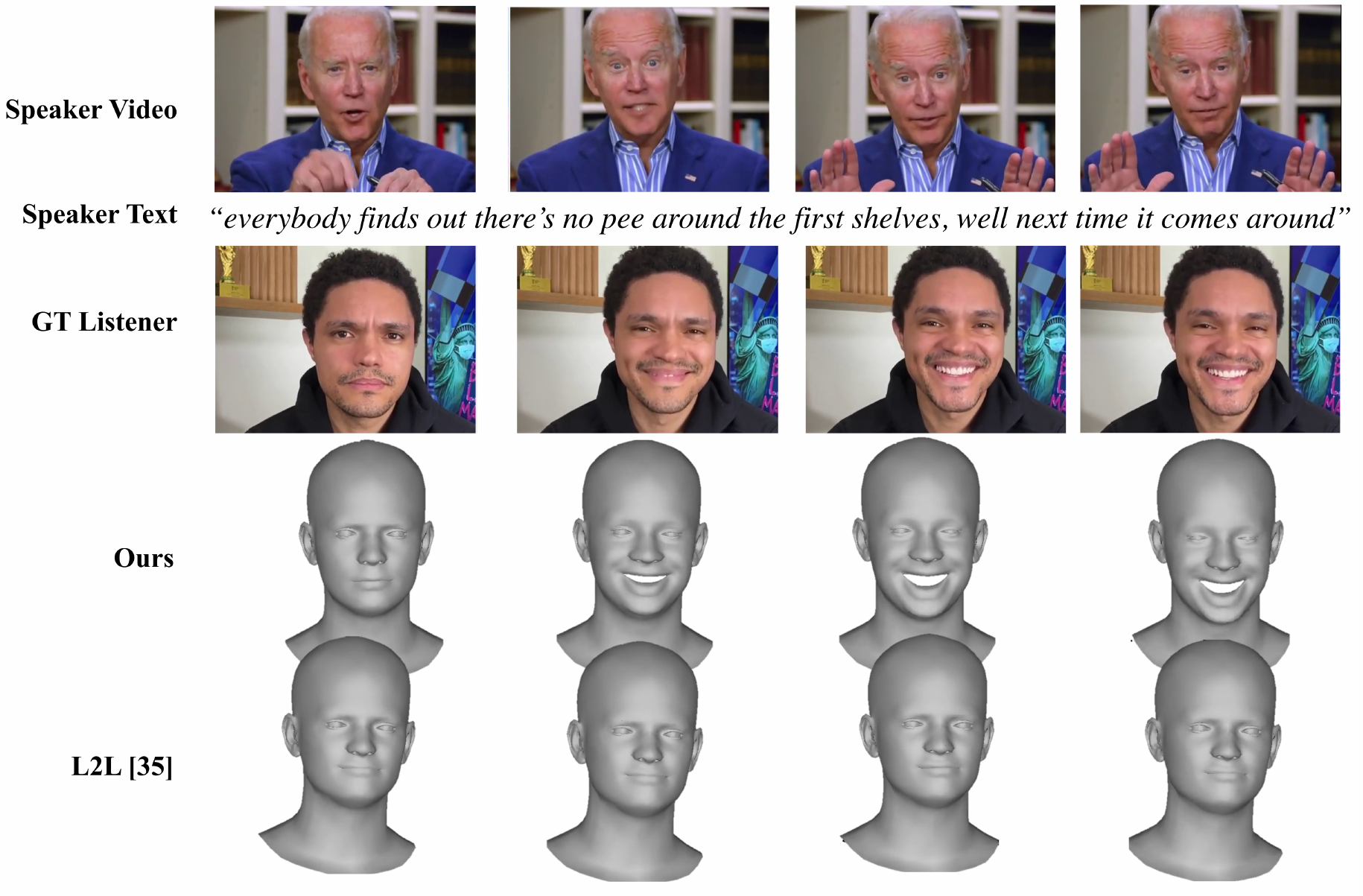}
    \caption{\textbf{Success in modeling humor.} In contrast to L2L~\cite{ng2022learning}, conditioned on speaker motion and speech, our text-based model correctly generates laughter in response to humorous language.}
    \label{fig:success}
\end{center}
\end{figure}

\subsection{Temporal Synchrony with the Speaker}
~\label{sec:temporal-analysis}
To understand which parts of our model contribute to its temporal performance, we evaluate variants of our model in Table~\ref{tab:ablation}. First, we consider the possibility that the temporal alignment and interleaving of language tokens (past listener and speaker text tokens) with listener motion tokens result in more temporally-synchronous motion. \textbf{Unaligned} is a variant we train with all text tokens prepended together \textit{in order}, followed by the autoregressively generated motion tokens. For \textbf{Unaligned}, we also remove the space before each VQ token. While \textbf{Full} performs slightly better than \textbf{Unaligned} in all metrics, the improvements are not significant. This indicates that the network can reason about the temporal interleaving of motion and language tokens without relying on their proximity in the input. 
Yet, we get notably poorer performance when we test the importance of word order by randomizing the input text tokens \textbf{Scrambled}. This suggests that our model leverages text \textit{ordering} as an important signal. 

We find evidence in the ground truth data that temporal information, essential for the synchrony of a dyad, can be learned from a text-only model. Analysis of the ground truth listener in the dataset demonstrates that punctuation is essential in regulating when to nod. For instance, around 51\% of the statements immediately before nods include some punctuation, while in the non-nodding case, only 15\% of utterances do. Similarly, smiles correlate more with ``!!" punctuation than plain faces (see 
Appendix~\ref{app:analysis} 
for analysis~). Therefore, we test the importance of punctuation in our model via two variations that introduce a new, fixed text token. We then replace all the input text with this new text token \textbf{FixTok} or replace all but punctuation \textbf{FixTok-Punc}. Note that in both these setups, we still preserve space tokens since we place spaces before each VQ token. We use a new text token instead of an existing one to avoid potential biases induced by pre-existing words or spaces. Comparing the two in Table~\ref{tab:results}, we see that punctuation significantly improves  speaker-listener synchrony via P-FD.

We conclude that through the formalization of sentence structure, speaker text transcriptions contain some temporal signal reflecting the nodding beat-like motions of the speaker, thus providing a hint for when it is appropriate for a listener to respond. This is also evident from cases where the appropriate response could rely on \textit{either} speaker motion or sentence structure as shown in Figure~\ref{fig:win-lose} (right).

\subsection{Semantically-appropriate Responses}
~\label{sec:semantic-analysis}
In the ground truth dataset, we analytically demonstrate common patterns from a listener's reaction to the speaker's words.  Figure~\ref{fig:lsaffect} plots the listener's facial affect ($1.0$ corresponds to very happy, $-1.0$ corresponds to very upset) associated with the top 100 most positive or negative phrases within the dataset. We calculate the average facial affect for each phrase during the utterance and in the 2 seconds that follow. The graphs exhibit a strong correlation between positive phrases and positive listener facial reactions and vice versa. Furthermore, Figure~\ref{fig:lsaffect} demonstrates that our finetuned model can capture the distribution of these associations well. Figure~\ref{fig:success} shows an example where our model generates laughter in response to humor--a joke that goes over the head of a motion-and-audio conditioned baseline. In Figure~\ref{fig:win-lose} (left), we show a case where our predicted listener is appropriately serious, despite the nervous laughter of the speaker.

Most notably, the fixed token experiment confirms our approach can properly model conversations' semantic alignment. Note that the fixed token models are the only ones with no semantic information since we replace all words with the same token. As a result, both models perform significantly worse than even \textbf{NoPT}. We further analyze semantic knowledge captured via word clouds in 
Appendix Figure~\ref{fig:wordcloud}.

\subsection{The Effect of Historical Context}
~\label{sec:history-analysis}
We consider the importance of historical lexical semantic context in conversation. A listener's response often depends on things said in the recent past. Figure~\ref{fig:history} demonstrates having no context at all results in worse performance. As we increase the amount of history we feed into the network, the predicted affect becomes more aligned with the ground truth, with a sweet spot of 8 seconds of history. However, adding too much context again results in poor performance.

\subsection{Limitations}

Given that listening is an inherently multimodal task involving visual and auditory signals from the speaker, our method is limited in that it does not take visual or audio input. For instance, Figure~\ref{fig:win-lose} (middle) shows an example where the speaker laughs, but the text does not contain an explicit joke. In other cases, the speaker may prompt the listener to nod through their motion or prosody rather than through words. More powerful language models may also improve our results. For instance, when prompted with humorous text, our model does not always generate laughter. Larger language models have demonstrated an improved capacity to model jokes~\cite{palm}, so integrating them into our framework may improve responses to such examples.


\section{Discussion}
We presented a transfer-based approach from pretrained large language models to human conversational gestures in dyadic interactions. This approach relies on the insight that gesture can be discretized into its atomic elements and treated as novel language tokens. We can, therefore, seamlessly integrate language and motion to extend state-of-the-art methods in language modeling to this setting. Integrating text input with other modalities for this task is a compelling direction for future work. 


%% file: supp.tex
\section{Results Video} \label{app:vid}
The supplementary video shows sequences of various individuals in different conversational settings from the Learning to Listen~\cite{ng2022learning} dataset. Below, we denote the time stamp range associated with the discussion - (@mm:ss-mm:ss).

The results show that our model successfully models semantically meaningful events. For instance, in a sequence where the speaker is telling a long joke, our model generates an appropriately timed laugh at the end of the joke (@00:45-00:56). Meanwhile, prior SOTA methods that rely on audio and motion do not capture these semantics and fail to generate any laugh (@00:58-01:10). In complex cases where the speaker's face mismatches the semantics of the text, e.g.~telling a joke while serious or smiling while saying something serious (@01:11-01:24), our method generates appropriate facial expressions. While the beat of the conversation is typically associated with prosody, we surprisingly demonstrate our model's ability to generate well-timed expressions \emph{and nods} from text alone (@01:25-01:38). Our main failure cases are long-tail cases when motion and audio are actually crucial in the semantics (@01:42-01:53). Note that our facial gestures do not just stay static throughout the entire sequence, but rather transitions in accordance to the
speaker’s small jokes or more serious comments.

Our method consistently generates more synchronous and realistic listener motion compared against \textbf{NN}, \textbf{Uncond}, and prior SOTA \textbf{L2L~\cite{ng2022learning}} (@02:29-02:55). See discussion in main paper and Table 1. While these baselines generate realistic looking motion, they fail to produce synchronous motion that match the mood of the conversation.

Furthermore, we show our method can generate multiple plausible listener trajectories conditioned on a single speaker sequence (@02:59-03:10), can generalize across various listeners (@03:13-03:33)\footnote{We use person-specific VQ models, since we found previously that this works better with the Vid2Vid visualization method described in Appendix~\ref{app:method}}, and can be used on arbitrary input text from a user (@03:40-@04:23).

\begin{figure}
    \centering
    \includegraphics[width=0.45\textwidth]{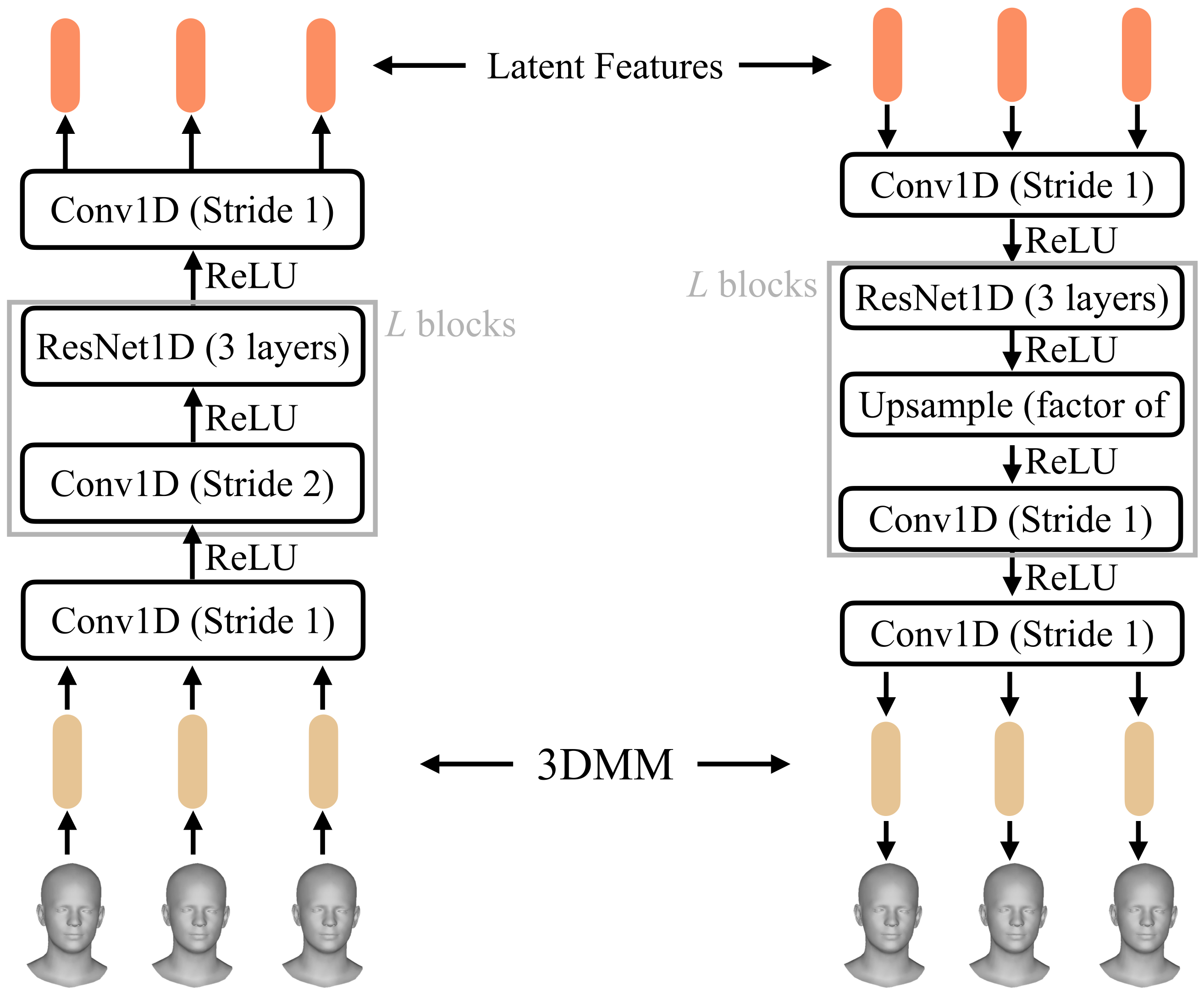}
    \caption{\textbf{Listener VQ-VAE.} Encoder (left) quantizes the listener motion, while the decoder (right) projects the learned discrete codebook tokens back into continuous motion space.}
    \label{fig:vqvae}
\end{figure}

\section{Method}
\label{app:method}
\paragraph{Implementation Details}
For our VQVAE, we use a codebook size of 256 and set $L=3$ for the number of downsampling layers in the encoder, so each VQ token corresponds to $r=2^3 = 8$ frames. When training the VQVAE, we use segments of $32$ frames. We use a learning rate of $2e-4$, train for $300,000$ steps, use $1000$ warmup steps, decay the learning rate by $0.05$ when we reach $200,000$ steps, and use the final model checkpoint. We use the following coefficients for the VQVAE training loss: $0.5$ for the velocity loss, $0.02$ for the embed loss, and $1$ for the reconstruction loss. The L2 error induced by quantization is marginal (0.08). 

For our motion prediction model, we use GPT2-Medium, which has 24 layers and 345M parameters. We train and evaluate our motion prediction model with at most $240$ frames (8 seconds) motion sequences. The maximum number of tokens in the input to the transformer is $480$, and we truncate the list of words from the text history when necessary to satisfy this constraint. During training, as in~\cite{t2mgpt}, we corrupt motion tokens in the input to the predictor with probability $0.5$. We use a learning rate of $5e-5$, train for a maximum of $50,000$ gradient steps, and terminate training early if the average training loss over $600$ gradient steps does not decrease by at least $0.001$. We use the last model checkpoint.

\paragraph{VQ-VAE Details} 
We train a VQ-VAE to quantize listener motion patches which serve as the input/output to the predictor model. Figure~\ref{fig:vqvae} show the internal details of the transformer-based VQ-VAE.

\paragraph{Sentiment Analysis Implementation} 
For analysis, we automatically predict affects using SOTA pretrained models for facial valence prediction. 
To calculate the facial sentiment, we use the pre-trained EMOCA~[14] sentiment classifier that was trained on AffectNet. Given 3DMM expression, pose, and shape coefficients, the EMOCA sentiment model is trained to predict the valence $\in [-1.0, 1.0]$, where $-1.0$ corresponds to unhappy/negative and $1.0$ corresponds to happy/positive facial expressions. To calculate the affect over a temporal window, we average the predicted valence for each frame in the sequence.

To calculate the text sentiment, we use a pre-trained SiBERT~\cite{hartmann2022emotionenglish}, which was fine-tuned from RoBERTa [Liu et.al.~2018]. The model outputs a binary classification label for positive (1) or negative (0) sentiment along with confidence scores. To calculate the valence, we verify the confidence is roughly correlated with text sentiment. Hence, if the predicted class label is negative, we negate the confidence value, else we take the confidence at face value. To compute the valence over a phrase, we can pass the full phrase into the model, which outputs a single confidence value and class label.

\begin{table}
\begin{center}
\begin{tabular}{@{}lrr@{}}
\toprule
  & Median Motion Length & Median Text Length \\
  \cmidrule{2-3}
  Trevor & 8 s & 31 words \\
  Conan & 3.2 s & 15.5 words \\
  Stephen & 2.7 s & 15 words \\

 \bottomrule
\end{tabular}
 \caption{\textbf{Median sequence lengths of validation sets.} For data preprocessing, we keep sequences where listeners are listening and crop out segments where the listener is either not visible, or are speaking. Then we divide the resulting data into segments of at most 8 seconds for training and evaluation. The table shows the median length of segments (in terms of seconds and number of words spoken by the speaker) in the remaining segments. We see that Trevor contains the longest sequences, while Conan and Stephen have considerably shorter segments.}
 \label{tab:times}
\end{center}
\end{table}

\paragraph{Visualization}
To improve the visual perceptibility of our results, we also train a person-specific 3DMM-to-video translation network.
We adopt the state-of-the-art video-to-video synthesis method Vid2Vid [Wang et.al.~Neurips 2018] to translate the grey scale 3DMM rendering results into full frames of a photo-realistic target video, in which the target listener mimics the facial expression and the head motion of the grey scale visualization. Our network learns to simulate the static background and the entire listener, where the face region is conditioned on the 3DMM rendering, while other components, such as hair and torso are compiled with the head pose. During the training session, the training data is extracted from a single video clip. The ground truth targets are the frames of the listener in the video, and the sources are the renderings of the corresponding 3DMM predictions.  

\begin{table*}\centering \footnotesize
\setlength{\tabcolsep}{5.0pt}.
\ra{1.3} 
\begin{tabular}{@{}lrrrcrrrrrrr@{}}
\toprule 
& PT GPT & text type & align && L2 $\downarrow$ & FD $\downarrow$ & variation & diversity & P-FD $\downarrow$ & L2 Affect ($10^2$) $\downarrow$ \\
\cmidrule{2-4} \cmidrule{6-11}
\rowcolor{light-gray}
\textit{GT} & & & & && & $0.18 \pm 0.01$ &  $3.60 \pm 0.11$ & & \\
\rowcolor{Gray}
Full & \cmark & given & \cmark && $\mathbf{0.81} \pm 0.03$ & $\mathbf{35.75} \pm 1.39$ & $0.15 \pm 0.01$ & $3.43 \pm 0.09$ & $\mathbf{39.09} \pm 1.39$ & $16.31 \pm 1.50$ \\
Random Train & \multicolumn{3}{c}{} && $1.05 \pm 0.04$ & $51.44 \pm 1.93$ & $0.17 \pm 0.02$ & $3.16 \pm 0.15$ & $53.98 \pm 1.95$ & $17.71 \pm 1.98$ \\
Random VQ & \multicolumn{3}{c}{} && $1.01 \pm 0.02$ & $44.16 \pm 1.19$ & $0.25 \pm 0.01$ & $4.68 \pm 0.09$ & $48.43 \pm 1.16$ & $\mathbf{14.81} \pm 1.28$ \\
NN & \multicolumn{3}{c}{} && $1.01 \pm 0.03$ & $48.43 \pm 1.83$ & $0.16 \pm 0.01$ & $3.17 \pm 0.12$ & $51.28 \pm 1.80$ & $20.48 \pm 2.06$ \\
Uncond & \multicolumn{3}{c}{} && $1.13 \pm 0.04$ & $58.56 \pm 2.00$ & $0.06\pm	0.00$ & $2.24	\pm0.02$ & $60.89	\pm2.02$ & $18.55	\pm1.99$\\
\cmidrule{2-4} \cmidrule{6-11}
NoPT  & \xmark & given & \cmark && $0.88	\pm0.02$ & $38.41	\pm1.25$ & $0.19	\pm0.01$ & $3.90	\pm0.09$ & $42.06	\pm1.25$ & $19.49	\pm2.00$ \\
Unaligned  & \cmark & given & \xmark && $0.84	\pm0.03$ & $36.59	\pm1.44$ & $0.16	\pm0.01$ & $3.59	\pm0.09$ & $40.19	\pm1.43$ & $16.99	\pm1.65$ \\
FixTok  & \cmark & fixed & \cmark && $0.95	\pm0.03$ & $45.25	\pm1.66$ & $0.12	\pm0.01$ & $3.03	\pm0.09$ & $48.22	\pm1.64$ & $21.60	\pm2.09$ \\
FixTok-Punc  & \cmark & punc.~+fixed & \cmark && $0.90	\pm0.03$ & $41.96	\pm1.51$ & $0.14	\pm0.01$ & $3.21	\pm0.10$ & $45.32	\pm1.50$ & $17.97	\pm2.02$ \\
\bottomrule
\end{tabular}
\caption{\textbf{Results on Conan dataset with Listener-agnostic VQ.} Each metric is averaged over the test set instances. Standard error is computed via bootstrap (using 10,000 samples).}
\label{tab:conan}
\end{table*}

\begin{table*}\centering \footnotesize
\setlength{\tabcolsep}{5.0pt}.
\ra{1.3} 
\begin{tabular}{@{}lrrrcrrrrrrr@{}}
\toprule 
& PT GPT & text type & align && L2 $\downarrow$ & FD $\downarrow$ & variation & diversity & P-FD $\downarrow$ & L2 Affect ($10^2$) $\downarrow$ \\
\cmidrule{2-4} \cmidrule{6-11}
\rowcolor{light-gray}
\textit{GT} & & & & && & $0.14	\pm0.01$ &  $3.11	\pm0.09$ & & \\
\rowcolor{Gray}
Full & \cmark & given & \cmark && $0.76	\pm0.02$ & $34.61	\pm1.25$ & $0.13	\pm0.01$ & $3.16	\pm0.07$ & $37.22	\pm1.24$ & $8.82	\pm0.72$\\
Random Train & \multicolumn{3}{c}{} && $0.95	\pm0.03$ & $47.08	\pm1.67$ & $0.10	\pm0.01$ & $2.54	\pm0.08$ & $48.82	\pm1.66$ & $10.46	\pm0.96$ \\
Random VQ & \multicolumn{3}{c}{} && $1.11	\pm0.02$ & $51.00	\pm1.28$ & $0.28	\pm0.01$ & $5.01	\pm0.07$ & $54.07	\pm1.25$ & $14.91	\pm1.10$\\
NN & \multicolumn{3}{c}{} && $0.88	\pm0.03$ & $43.10	\pm1.56$ & $0.09	\pm0.01$ & $2.61	\pm0.08$ & $45.03	\pm1.54$ & $8.87	\pm0.81$\\
Uncond & \multicolumn{3}{c}{} && $1.62	\pm0.05$ & $85.23	\pm2.69$ & $0.09	\pm0.00$ & $2.67	\pm0.05$ & $87.31	\pm2.67$ & $37.62	\pm1.36$\\
\cmidrule{2-4} \cmidrule{6-11}
NoPT  & \xmark & given & \cmark && $0.78	\pm0.02$ & $36.18	\pm1.27$ & $0.13	\pm0.01$ & $3.16	\pm0.07$ & $38.55	\pm1.29$ & $\mathbf{8.35}	\pm0.64$ \\
Unaligned  & \cmark & given & \xmark && $\mathbf{0.73}	\pm0.02$ & $\mathbf{33.47}	\pm1.21$ & $0.12	\pm0.00$ & $3.14	\pm0.07$ & $\mathbf{35.90}	\pm1.21$ & $8.45	\pm0.66$ \\
FixTok  & \cmark & fixed & \cmark && $0.84	\pm0.02$ & $41.04	\pm1.35$ & $0.09	\pm0.01$ & $2.63	\pm0.07$ & $43.14	\pm1.35$ & $10.74	\pm0.90$ \\
FixTok-Punc  & \cmark & punc.~+fixed & \cmark && $0.83	\pm0.02$ & $39.80	\pm1.26$ & $0.11	\pm0.00$ & $2.86	\pm0.07$ & $42.08	\pm1.26$ & $10.40	\pm0.83$ \\
\bottomrule
\end{tabular}
\caption{\textbf{Results on Stephen dataset with Listener-agnostic VQ.} Each metric is averaged over the test set instances. Standard error is computed via bootstrap (using 10,000 samples).}
\label{tab:stephen}
\end{table*}

\begin{table}
\begin{center}
\begin{tabular}{@{}lrrr@{}}
\toprule
  & small nod & big nod & plain face \\
\cmidrule{2-4}
,         & 31\% & 38\% & 29\% \\
. / ! / ? & 51\% & 44\% & 15\% \\
?         & 3\% & 13\% & 0\% \\ 
!         & 24\% & 13\% & 1\% \\ 
...       & 5\% & 0\% & 1\% \\
``and" / ``like" / ``or" & 9\% & 0\% & 8\% \\
 \bottomrule
\end{tabular}
 \caption{\textbf{Nods and Punctuation} We calculate how often punctuation or conjunctions occur in the tokens leading up to a small/big/no nod sequence. There is strong evidence that nods are often associated with end punctuation. It is therefore possible for our text-only model to learn appropriate timing of motion that is often usually associated with prosody.}
 \label{tab:punc}
\end{center}
\end{table}

\begin{table*}\centering \footnotesize
\setlength{\tabcolsep}{12.0pt}.
\ra{1.3} 
\begin{tabular}{@{}lrrrrrr@{}}
\toprule 
& L2 $\downarrow$ & FD $\downarrow$ & variation & diversity & P-FD $\downarrow$ & L2 Affect ($10^2$) $\downarrow$ \\
\cmidrule{2-7}
\rowcolor{light-gray}
\textit{GT} & & & $0.11$ &  $2.59$ & & \\
hist0 & $0.44 \pm 0.02$ & $19.47 \pm 0.8$ & $0.099 \pm 0.005$ & $2.53 \pm 0.06$ & $20.74 \pm 0.8$ & $6.84 \pm 0.50$ \\
hist3 (Ours) & $0.43 \pm 0.02$ & $18.22 \pm 0.7$ & $0.116 \pm 0.005$ & $2.81 \pm 0.06$ & $19.63 \pm 0.8$ & $6.36 \pm 0.47$ \\
hist8 & $0.44 \pm 0.02$ & $18.32 \pm 0.7$ & $0.124 \pm 0.005$ & $2.87 \pm 0.06$ & $19.76 \pm 0.7$ & $6.19 \pm 0.46$\\
histFull & $0.48 \pm 0.02$ & $19.71 \pm 0.9$ & $0.136 \pm 0.005$ & $3.07 \pm 0.06$ & $21.24 \pm 0.9$ & $6.86 \pm 0.47$\\
\bottomrule
\end{tabular}
\caption{\textbf{Effect of textual history.} Comparison against ground truth annotations (GT). $\downarrow$ indicates lower is better; for no arrow, closer to GT is better. Each metric is averaged over the test set instances. Standard error is computed via bootstrap (using 10,000 samples).}
\label{tab:history}
\end{table*}

\section{Results}
\label{app:results}
\paragraph{Data processing} We process videos for two other listeners in the L2L dataset~\cite{ng2022learning}: Conan O'Brien and Stephen Colbert. As shown in Table~\ref{tab:times}, these listeners are shown listening in the videos far less frequently than Trevor Noah. Consequently, the average length of motion segments and the average number of text tokens per segment in our processed data is lower for these listeners than for Trevor Noah. We also noticed several cases in which the listener interrupts the speaker in these videos, so we modified the preprocessing to filter out segments in which the diarization says that the listener speaks. In order to improve the quality of the test set, we manually viewed the segments in the test set and removed segments if (1) the EMOCA pseudo-ground truth is noisy, (2) the listener said more words than the speaker, or (3) the listener was not shown or the clip was not part of a two-person conversation involving the listener.

\paragraph{Individual Styles of Listening Motion}
The decision to train different models per-person comes from the idea that each person has a characteristic way of listening. By training person-specific models, we can capture these more fine-grain characteristic details of listener motion, which is difficult to do when considering all identities at once. Nevertheless, our experiments demonstrate that
\emph{the text-based generation approach continues to outperform baselines when trained on different listeners} in the dataset (e.g.~Conan, Stephen). For simplicity, we report on Trevor Noah in the main paper since the dataset is the cleanest and has the greatest amount of long, uninterrupted conversational video clips ($\geq 8$s) where both the speaker and listener are visible. 

\paragraph{Nearest-neighbor and Random Retrieval Baselines}
As described in the main paper, we our nearest neighbor baseline uses sentence embeddings~\cite{reimers-2019-sentence-bert} to retrieve a motion sequence from the training data for each evaluation sequence. In particular, we consider all training sequences that have a length of 8 seconds and associate each such sequence with the words that are in the span from 3 seconds before the start of the sequence to the end of the sequence.\footnote{Each unique sequence of text tokens is associated with a single motion sequence, so when the same text occurs for multiple training sequences, we drop all but one training sequence.} The random retrieval baseline (``Random Train'') uses the same segments from training as the nearest neighbor baseline to retrieve a motion sequence (but, unlike the nearest neighbor baseline, it ignores the text and instead selects one of the training sequences uniformly at random). In both the nearest neighbor and the random retrieval baselines, since evaluation sequences may be shorter than 8 seconds, we truncate the retrieved training sequence so that it has the same length as the target evaluation sequence.

\paragraph{Results with a Listener-Agnostic VQ}
Here we report results on two other listeners, Conan and Stephen, from the L2L dataset~\cite{ng2022learning}. For these experiments, we use a ``global'' VQ-VAE trained across all three listeners (Trevor, Conan, and Stephen) and train separate motion prediction transformers for each listener. Given the noisiness of the Conan dataset, we even found this listener-agnostic approach outperforms training a listener-specific VQ-VAE on the Conan validation set.
Table~\ref{tab:stephen} and~\ref{tab:conan} contain the full quantitative results on the Stephen and Conan held-out test sets respectively.
\emph{\textbf{Even in a listener agnostic setup, our text-based generation approach outperforms existing baselines.}} Several trends observed on the results for Trevor also apply to these other listeners. In most metrics, the text-based generation approach outperforms the random and nearest-neighbor baselines. The \textbf{Uncond.} baseline and ablations in which a fixed text token is used in place of the true text tokens, \textbf{FixTok} and \textbf{FixTok-Punc}, all perform worse than in cases where the text tokens are given, confirming that our network successfully leverages semantics. Furthermore, \textbf{FixTok} performs worse than \textbf{FixTok-Punc}, providing evidence that punctuation is essential in listener synchrony. Finally, aligning the text tokens with the VQ tokens according to the text token timestamps does not make a significant difference. 

There are also some differences in the results. For instance, the pre-trained GPT2 initialization matters less and less on shorter and shorter segments (matters less on Conan and lesser on Stephen, with Stephen being the shortest segments). One factor that may contribute to this difference is that there are far fewer text tokens (and fewer punctuation characters) per segment in the Conan/Stephen segments compared to the Trevor segments. Since pre-training is likely to help most significantly in the processing of the text, with fewer text tokens, the effect of pre-training is less significant. 

\paragraph{Learning to Listen (L2L)~\cite{ng2022learning} Implementation Details} To implement this baseline, we take the pre-trained, person-specific model that corresponds to Trevor Noah. However, in order to remove the bias of EMOCA vs.~DECA, we visualize the mesh of the L2L method with the same texturing and shape taken from our updated EMOCA annotations. As a result, we make sure to visualize the meshes using the same level of detail and upgraded visualization method to remove the bias that AMT evaluators would prefer our method over L2L due to added details.

\noindent \textbf{Qualitative Evaluation Details.}
For each Ours vs.~NN motion, vs. unconditioned, vs.~L2L, vs.~GT, we generate 50 A-B tests. For each of the 50 A-B tests, we ask 3 different evaluators, totalling to 600 evaluators. For the NN, unconditioned, and GT tests, each A-B test contained 14 questions. For L2L, they contained 12 questions since we only took overlapping sequences from the test set between L2L and our method.
Prior to the actual test, we provide a headphone check to make sure the evaluators are listening to audio. However, we do not ask additional questions that check to see if they are actually listening to the speech. The landing page describes the task and walks evaluators through 2 examples. 
To ensure the evaluators are not just randomly clicking, we include 3 questions with an obvious mismatch (one speaker laughing while the listener is neutral) twice. If the evaluator selects a different response for these duplicated questions, we do not allow them to submit. 

\section{Analysis of the Text-based Method}
\label{app:analysis}

\paragraph{Punctuation and Temporal Synchrony}
In Table~\ref{tab:punc}, we analyze evidence that the network can learn temporal synchrony of movements, primarily associated with prosody, from text alone. We utilize the trained VQ codebook to find motion tokens that often are associated with nod movements. Using the psudo ground truth annotations across all the datasets, we convert the motion sequences into quantized codebook indices. We then find all instances of the ``nodding" VQ tokens in the dataset. For each found ``nodding" token, we take the 5 text tokens right before the ``nodding" token. This corresponds to finding which text tokens lead to a nod. Within these 5 text tokens, we count how often a certain punctuation or conjunction occurs. We discover that in the ground truth data, there is strong correlation between punctuation and nods. We show our model similarly relies on punctuation in Table 2 with \textbf{FixedTok-Punc}, where peformance worsens when we remove all forms of punctuation \textbf{FixedTok}.

\begin{figure}[t]
\begin{center}
\includegraphics[width=\linewidth]{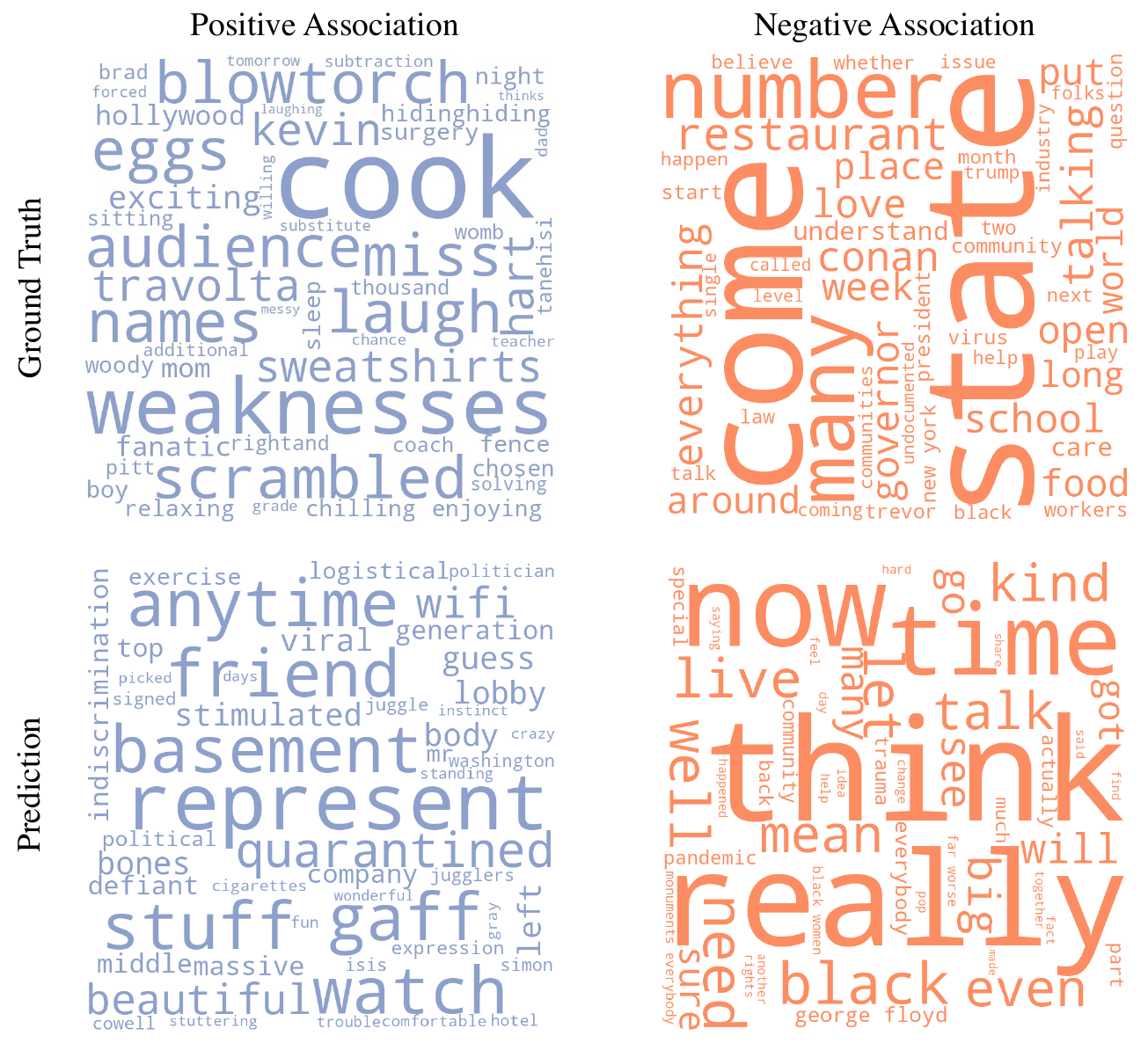}
\caption{\textbf{Positive negative word clouds.} We fetch 100 listener sequences corresponding to the most positive (left) and negative (right) affect, respectively. The word cloud exhibits all words occurring two seconds before the expression. Both ground truth (top) and our predictions (bottom) capture positive (exercise, quarantined, mom) and negative (workers, pandemic, George Floyd) sentiments reflective of the Covid era during which the dataset was collected.}
\label{fig:wordcloud}
\end{center}
\end{figure}

\paragraph{Word Clouds}
In Figure~\ref{fig:wordcloud}, we analyze the semantic knowledge captured by our model by visualizing word-sentiment associations. We retrieve the most positive (left) and negative (right) facial expression sequences from both ground truth (top) and our predictions (bottom) and visualize the words from phrases that precede the expression. From the word clouds, we see that our model picks up on common topics often associated with positive (watch, defiant, viral, friend) and negative (George Floyd, pandemic, trauma) sentiments from the 2020 Covid pandemic era.

\paragraph{Effect of Historical Context} 
In Figure 5 of the main paper, we show the effect of increasing historical context on the performance of the model. In Table~\ref{tab:history}, we show the full metrics computed for each experiment. Generally, adding more history improves performance up to a certain point. After 8 seconds of history, the model's performance starts to plateau and performance worsens across all metrics when too much history is given. 

\section{Addendum: Content Overlap Between Videos}
We found that there are some original YouTube videos that are excerpts of other full-interview videos. Therefore, there are videos in the test sets whose content (though not identity) overlaps with videos in the training sets. We identified this overlap via text n-gram overlap between transcripts. In particular, in the Trevor Noah test set, there are 105 clips in the test set that come from a video which overlaps with two videos in the training set (8/47 of the videos used in the AMT comparisons come from this video). In the Conan test set, there are 8 clips that come from videos which overlap with videos in the training set.
On the project’s GitHub page, we have provided updated data files omitting these clips from the respective test sets.

\noindent
For the methods for which we have the required checkpoints (random train, random VQ, nearest-neighbor, and text-conditioned LM), we reran the automated evaluations in Table~\ref{tab:results} with the updated Trevor Noah test set, and the results are in Table~\ref{tab:results-no-overlap}. As in Table~\ref{tab:results}, the text-conditioned LM presented here remains the best-performing among the random train, random VQ, nearest-neighbor, and text-conditioned LM approaches.
\begin{table*}\centering \footnotesize
\setlength{\tabcolsep}{12.0pt}.
\ra{1.3} 
\begin{tabular}{@{}lrrrrrr@{}}
\toprule 
& L2 $\downarrow$ & FD $\downarrow$ & variation & diversity & P-FD $\downarrow$ & L2 Affect ($10^2$) $\downarrow$ \\
\cmidrule{2-7}
\rowcolor{light-gray}
\textit{GT} & & & $0.12$ &  $2.59$ & & \\
Random Train & $0.67 \pm 0.02$ & $32.43 \pm 1.2$ & $0.094 \pm 0.0057$ & $2.34 \pm 0.07 $& $33.57 \pm 1.21$ & $10.44 \pm 0.74$ \\
Random VQ & $0.71 \pm 0.01$ & $29.30 \pm 0.74$ & $0.26 \pm 0.0053$ & $4.79 \pm 0.06$ & $31.42 \pm 0.75$ & $10.04 \pm 0.56$ \\
NN & $0.61 \pm 0.02$ & $28.32 \pm 1.16$ & $0.086 \pm 0.0049$ & $2.25 \pm 0.06$ & $29.57 \pm 1.18$ & $9.29 \pm 0.66$ \\
\rowcolor{Gray}
Full & $0.53 \pm 0.02$ & $22.13 \pm 0.88$ & $0.13 \pm 0.0054$ & $3.00 \pm 0.07$ & $23.83 \pm 0.90$ & $7.77 \pm 0.62$ \\
\bottomrule
\end{tabular}
\caption{\textbf{Results on Test Set without Overlapping Clips.} Comparison against ground truth annotations (GT). $\downarrow$ indicates lower is better; closer to GT is better for no arrow. We average each metric over the test set instances. Standard error is computed via bootstrap (using 10,000 samples).}
\label{tab:results-no-overlap}
\end{table*}


%% file: main.bbl
\begin{thebibliography}{10}\itemsep=-1pt

\bibitem{ahuja2019react}
Chaitanya Ahuja, Shugao Ma, Louis-Philippe Morency, and Yaser Sheikh.
\newblock To react or not to react: End-to-end visual pose forecasting for
  personalized avatar during dyadic conversations.
\newblock In {\em 2019 International Conference on Multimodal Interaction},
  pages 74--84, 2019.

\bibitem{blanz1999morphable}
Volker Blanz and Thomas Vetter.
\newblock A morphable model for the synthesis of 3d faces.
\newblock In {\em Proceedings of the 26th annual conference on Computer
  graphics and interactive techniques}, pages 187--194, 1999.

\bibitem{Bohus_Horvitz_2010}
Dan Bohus and Eric Horvitz.
\newblock Facilitating multiparty dialog with gaze, gesture, and speech.
\newblock In {\em International Conference on Multimodal Interfaces and the
  Workshop on Machine Learning for Multimodal Interaction}, ICMI-MLMI '10, New
  York, NY, USA, 2010. Association for Computing Machinery.

\bibitem{pyannote2}
Herv{\'e} {Bredin} and Antoine {Laurent}.
\newblock {End-to-end speaker segmentation for overlap-aware resegmentation}.
\newblock In {\em Proc. Interspeech 2021}, 2021.

\bibitem{pyannote1}
Herv{\'e} {Bredin}, Ruiqing {Yin}, Juan~Manuel {Coria}, Gregory {Gelly}, Pavel
  {Korshunov}, Marvin {Lavechin}, Diego {Fustes}, Hadrien {Titeux}, Wassim
  {Bouaziz}, and Marie-Philippe {Gill}.
\newblock {pyannote.audio: neural building blocks for speaker diarization}.
\newblock In {\em ICASSP 2020, IEEE International Conference on Acoustics,
  Speech, and Signal Processing}, 2020.

\bibitem{brown2020language}
Tom Brown, Benjamin Mann, Nick Ryder, Melanie Subbiah, Jared~D Kaplan, Prafulla
  Dhariwal, Arvind Neelakantan, Pranav Shyam, Girish Sastry, Amanda Askell,
  et~al.
\newblock Language models are few-shot learners.
\newblock {\em Advances in neural information processing systems},
  33:1877--1901, 2020.

\bibitem{cao2013facewarehouse}
Chen Cao, Yanlin Weng, Shun Zhou, Yiying Tong, and Kun Zhou.
\newblock Facewarehouse: A 3d facial expression database for visual computing.
\newblock {\em IEEE Transactions on Visualization and Computer Graphics}, 2013.

\bibitem{Cassell1994}
Justine Cassell, Catherine Pelachaud, Norman Badler, Mark Steedman, Brett
  Achorn, Tripp Becket, Brett Douville, Scott Prevost, and Matthew Stone.
\newblock Animated conversation: Rule-based generation of facial expression,
  gesture and spoken intonation for multiple conversational agents.
\newblock In {\em Proceedings of the 21st Annual Conference on Computer
  Graphics and Interactive Techniques}, SIGGRAPH '94, page 413–420, New York,
  NY, USA, 1994. Association for Computing Machinery.

\bibitem{CassellThorisson1999}
Justine Cassell and Kristinn~R. Thorisson.
\newblock The power of a nod and a glance: Envelope vs. emotional feedback in
  animated conversational agents.
\newblock {\em Applied Artificial Intelligence}, 13(4-5):519--538, 1999.

\bibitem{chartrand1999chameleon}
Tanya~L Chartrand and John~A Bargh.
\newblock The chameleon effect: the perception--behavior link and social
  interaction.
\newblock {\em Journal of personality and social psychology}, 76(6):893, 1999.

\bibitem{palm}
Aakanksha Chowdhery, Sharan Narang, Jacob Devlin, Maarten Bosma, Gaurav Mishra,
  Adam Roberts, Paul Barham, Hyung~Won Chung, Charles Sutton, Sebastian
  Gehrmann, Parker Schuh, Kensen Shi, Sasha Tsvyashchenko, Joshua Maynez,
  Abhishek Rao, Parker Barnes, Yi Tay, Noam~M. Shazeer, Vinodkumar Prabhakaran,
  Emily Reif, Nan Du, Benton~C. Hutchinson, Reiner Pope, James Bradbury, Jacob
  Austin, Michael Isard, Guy Gur-Ari, Pengcheng Yin, Toju Duke, Anselm
  Levskaya, Sanjay Ghemawat, Sunipa Dev, Henryk Michalewski, Xavier Garc{\'i}a,
  Vedant Misra, Kevin Robinson, Liam Fedus, Denny Zhou, Daphne Ippolito, David
  Luan, Hyeontaek Lim, Barret Zoph, Alexander Spiridonov, Ryan Sepassi, David
  Dohan, Shivani Agrawal, Mark Omernick, Andrew~M. Dai,
  Thanumalayan~Sankaranarayana Pillai, Marie Pellat, Aitor Lewkowycz, Erica
  Moreira, Rewon Child, Oleksandr Polozov, Katherine Lee, Zongwei Zhou, Xuezhi
  Wang, Brennan Saeta, Mark D{\'i}az, Orhan Firat, Michele Catasta, Jason Wei,
  Kathleen~S. Meier-Hellstern, Douglas Eck, Jeff Dean, Slav Petrov, and Noah
  Fiedel.
\newblock Palm: Scaling language modeling with pathways.
\newblock {\em ArXiv}, abs/2204.02311, 2022.

\bibitem{chu2018face}
Hang Chu, Daiqing Li, and Sanja Fidler.
\newblock A face-to-face neural conversation model.
\newblock In {\em Proceedings of the IEEE Conference on Computer Vision and
  Pattern Recognition}, pages 7113--7121, 2018.

\bibitem{condon_ogston_1966}
W.~S. Condon and W.~D. Ogston.
\newblock Sound film analysis of normal and pathological behavior patterns.
\newblock {\em The Journal of Nervous and Mental Disease}, 143(4), 1966.

\bibitem{emoca}
Radek Danecek, Michael~J. Black, and Timo Bolkart.
\newblock {EMOCA}: {E}motion driven monocular face capture and animation.
\newblock In {\em Conference on Computer Vision and Pattern Recognition
  (CVPR)}, pages 20311--20322, 2022.

\bibitem{learn2smile2017}
Will Feng, Anitha Kannan, Georgia Gkioxari, and Larry Zitnick.
\newblock Learn2smile: Learning non-verbal interaction through observation.
\newblock {\em IROS}, 2017.

\bibitem{deca}
Yao Feng, Haiwen Feng, Michael~J. Black, and Timo Bolkart.
\newblock Learning an animatable detailed {3D} face model from in-the-wild
  images.
\newblock {\em ACM Transactions on Graphics (ToG), Proc. SIGGRAPH}, 40(8),
  2021.

\bibitem{emocav2}
Panagiotis~P. Filntisis, George Retsinas, Foivos Paraperas-Papantoniou,
  Athanasios Katsamanis, Anastasios Roussos, and Petros Maragos.
\newblock Visual speech-aware perceptual 3d facial expression reconstruction
  from videos.
\newblock {\em arXiv preprint arXiv:2207.11094}, 2022.

\bibitem{geng2023affective}
Scott Geng, Revant Teotia, Purva Tendulkar, Sachit Menon, and Carl Vondrick.
\newblock Affective faces for goal-driven dyadic communication.
\newblock {\em arXiv preprint arXiv:2301.10939}, 2023.

\bibitem{ginosar2019learning}
Shiry Ginosar, Amir Bar, Gefen Kohavi, Caroline Chan, Andrew Owens, and
  Jitendra Malik.
\newblock Learning individual styles of conversational gesture.
\newblock In {\em Proceedings of the IEEE/CVF Conference on Computer Vision and
  Pattern Recognition}, pages 3497--3506, 2019.

\bibitem{gratch2006virtual}
Jonathan Gratch, Anna Okhmatovskaia, Francois Lamothe, Stacy Marsella, Mathieu
  Morales, Rick~J van~der Werf, and Louis-Philippe Morency.
\newblock Virtual rapport.
\newblock In {\em International Workshop on Intelligent Virtual Agents}, pages
  14--27. Springer, 2006.

\bibitem{greenwood2017predicting}
David Greenwood, Stephen Laycock, and Iain Matthews.
\newblock Predicting head pose in dyadic conversation.
\newblock In {\em International Conference on Intelligent Virtual Agents},
  pages 160--169. Springer, 2017.

\bibitem{guo2022generating}
Chuan Guo, Shihao Zou, Xinxin Zuo, Sen Wang, Wei Ji, Xingyu Li, and Li Cheng.
\newblock Generating diverse and natural 3d human motions from text.
\newblock In {\em Proceedings of the IEEE/CVF Conference on Computer Vision and
  Pattern Recognition}, pages 5152--5161, 2022.

\bibitem{hartmann2022emotionenglish}
Jochen Hartmann.
\newblock Emotion english distilroberta-base, 2022.

\bibitem{heusel2017gans}
Martin Heusel, Hubert Ramsauer, Thomas Unterthiner, Bernhard Nessler, and Sepp
  Hochreiter.
\newblock Gans trained by a two time-scale update rule converge to a local nash
  equilibrium.
\newblock {\em arXiv preprint arXiv:1706.08500}, 2017.

\bibitem{huang2011virtual}
Lixing Huang, Louis-Philippe Morency, and Jonathan Gratch.
\newblock Virtual rapport 2.0.
\newblock In {\em International workshop on intelligent virtual agents}, pages
  68--79. Springer, 2011.

\bibitem{huang2017dyadgan}
Yuchi Huang and Saad~M Khan.
\newblock Dyadgan: Generating facial expressions in dyadic interactions.
\newblock In {\em Proceedings of the IEEE Conference on Computer Vision and
  Pattern Recognition Workshops}, pages 11--18, 2017.

\bibitem{jonell2019learning}
Patrik Jonell, Taras Kucherenko, Erik Ekstedt, and Jonas Beskow.
\newblock Learning non-verbal behavior for a social robot from youtube videos.
\newblock In {\em ICDL-EpiRob Workshop on Naturalistic Non-Verbal and Affective
  Human-Robot Interactions, Oslo, Norway, August 19, 2019}, 2019.

\bibitem{joo2019towards}
Hanbyul Joo, Tomas Simon, Mina Cikara, and Yaser Sheikh.
\newblock Towards social artificial intelligence: Nonverbal social signal
  prediction in a triadic interaction.
\newblock In {\em Proceedings of the IEEE/CVF Conference on Computer Vision and
  Pattern Recognition}, pages 10873--10883, 2019.

\bibitem{kendon1970movement}
Adam Kendon.
\newblock Movement coordination in social interaction: Some examples described.
\newblock {\em Acta psychologica}, 32:101--125, 1970.

\bibitem{kucherenko2020gesticulator}
Taras Kucherenko, Patrik Jonell, Sanne Van~Waveren, Gustav~Eje Henter, Simon
  Alexandersson, Iolanda Leite, and Hedvig Kjellstr{\"o}m.
\newblock Gesticulator: A framework for semantically-aware speech-driven
  gesture generation.
\newblock In {\em Proceedings of the 2020 International Conference on
  Multimodal Interaction}, pages 242--250, 2020.

\bibitem{LaFrance1979}
Marianne LaFrance.
\newblock Nonverbal synchrony and rapport: Analysis by the cross-lag panel
  technique.
\newblock {\em Social Psychology Quarterly}, 42(1):66--70, 1979.

\bibitem{FLAME:SiggraphAsia2017}
Tianye Li, Timo Bolkart, Michael.~J. Black, Hao Li, and Javier Romero.
\newblock Learning a model of facial shape and expression from {4D} scans.
\newblock {\em ACM Transactions on Graphics, (Proc. SIGGRAPH Asia)}, 36(6),
  2017.

\bibitem{liu2023language}
Hao Liu, Wilson Yan, and Pieter Abbeel.
\newblock Language quantized autoencoders: Towards unsupervised text-image
  alignment.
\newblock {\em arXiv preprint arXiv:2302.00902}, 2023.

\bibitem{lu2021pretrained}
Kevin Lu, Aditya Grover, Pieter Abbeel, and Igor Mordatch.
\newblock Pretrained transformers as universal computation engines.
\newblock {\em arXiv preprint arXiv:2103.05247}, 1, 2021.

\bibitem{ng2022learning}
Evonne Ng, Hanbyul Joo, Liwen Hu, Hao Li, Trevor Darrell, Angjoo Kanazawa, and
  Shiry Ginosar.
\newblock Learning to listen: Modeling non-deterministic dyadic facial motion.
\newblock In {\em Proceedings of the IEEE/CVF Conference on Computer Vision and
  Pattern Recognition}, pages 20395--20405, 2022.

\bibitem{nojavanasghari2018interactive}
Behnaz Nojavanasghari, Yuchi Huang, and Saad Khan.
\newblock Interactive generative adversarial networks for facial expression
  generation in dyadic interactions.
\newblock {\em arXiv preprint arXiv:1801.09092}, 2018.

\bibitem{paysan20093d}
Pascal Paysan, Reinhard Knothe, Brian Amberg, Sami Romdhani, and Thomas Vetter.
\newblock A 3d face model for pose and illumination invariant face recognition.
\newblock In {\em 2009 Sixth IEEE International Conference on Advanced Video
  and Signal Based Surveillance}, pages 296--301. Ieee, 2009.

\bibitem{petrovich2022temos}
Mathis Petrovich, Michael~J Black, and G{\"u}l Varol.
\newblock Temos: Generating diverse human motions from textual descriptions.
\newblock In {\em Computer Vision--ECCV 2022: 17th European Conference, Tel
  Aviv, Israel, October 23--27, 2022, Proceedings, Part XXII}, pages 480--497.
  Springer, 2022.

\bibitem{whisper}
Alec Radford, Jong~Wook Kim, Tao Xu, Greg Brockman, Christine McLeavey, and
  Ilya Sutskever.
\newblock Robust speech recognition via large-scale weak supervision.
\newblock {\em ArXiv}, abs/2212.04356, 2022.

\bibitem{gpt2}
Alec Radford, Jeff Wu, Rewon Child, David Luan, Dario Amodei, and Ilya
  Sutskever.
\newblock Language models are unsupervised multitask learners.
\newblock 2019.

\bibitem{reid2022can}
Machel Reid, Yutaro Yamada, and Shixiang~Shane Gu.
\newblock Can wikipedia help offline reinforcement learning?
\newblock {\em arXiv preprint arXiv:2201.12122}, 2022.

\bibitem{reimers-2019-sentence-bert}
Nils Reimers and Iryna Gurevych.
\newblock Sentence-bert: Sentence embeddings using siamese bert-networks.
\newblock In {\em Proceedings of the 2019 Conference on Empirical Methods in
  Natural Language Processing}. Association for Computational Linguistics, 11
  2019.

\bibitem{sonlu2021conversational}
Sinan Sonlu, U{\u{g}}ur G{\"u}d{\"u}kbay, and Funda Durupinar.
\newblock A conversational agent framework with multi-modal personality
  expression.
\newblock {\em ACM Transactions on Graphics (TOG)}, 40(1):1--16, 2021.

\bibitem{tevet2022motionclip}
Guy Tevet, Brian Gordon, Amir Hertz, Amit~H Bermano, and Daniel Cohen-Or.
\newblock Motionclip: Exposing human motion generation to clip space.
\newblock In {\em Computer Vision--ECCV 2022: 17th European Conference, Tel
  Aviv, Israel, October 23--27, 2022, Proceedings, Part XXII}, pages 358--374.
  Springer, 2022.

\bibitem{tevet2022human}
Guy Tevet, Sigal Raab, Brian Gordon, Yonatan Shafir, Daniel Cohen-Or, and
  Amit~H Bermano.
\newblock Human motion diffusion model.
\newblock {\em arXiv preprint arXiv:2209.14916}, 2022.

\bibitem{vqvae}
A{\"a}ron van~den Oord, Oriol Vinyals, and Koray Kavukcuoglu.
\newblock Neural discrete representation learning.
\newblock In {\em NIPS}, 2017.

\bibitem{transformer}
Ashish Vaswani, Noam~M. Shazeer, Niki Parmar, Jakob Uszkoreit, Llion Jones,
  Aidan~N. Gomez, Lukasz Kaiser, and Illia Polosukhin.
\newblock Attention is all you need.
\newblock {\em ArXiv}, abs/1706.03762, 2017.

\bibitem{t2mgpt}
Jianrong Zhang, Yangsong Zhang, Xiaodong Cun, Shaoli Huang, Yong Zhang, Hongwei
  Zhao, Hongtao Lu, and Xiaodong Shen.
\newblock T2m-gpt: Generating human motion from textual descriptions with
  discrete representations.
\newblock {\em ArXiv}, abs/2301.06052, 2023.

\bibitem{zhang2023generating}
Jianrong Zhang, Yangsong Zhang, Xiaodong Cun, Shaoli Huang, Yong Zhang, Hongwei
  Zhao, Hongtao Lu, and Xi Shen.
\newblock T2m-gpt: Generating human motion from textual descriptions with
  discrete representations.
\newblock In {\em IEEE Conf. Comput. Vis. Pattern Recog.}, 2023.

\bibitem{zhou2022responsive}
Mohan Zhou, Yalong Bai, Wei Zhang, Ting Yao, Tiejun Zhao, and Tao Mei.
\newblock Responsive listening head generation: A benchmark dataset and
  baseline.
\newblock In {\em ECCV}, 2022.

\bibitem{zollhofer2018state}
Michael Zollh{\"o}fer, Justus Thies, Pablo Garrido, Derek Bradley, Thabo
  Beeler, Patrick P{\'e}rez, Marc Stamminger, Matthias Nie{\ss}ner, and
  Christian Theobalt.
\newblock State of the art on monocular 3d face reconstruction, tracking, and
  applications.
\newblock In {\em Computer Graphics Forum}, 2018.

\end{thebibliography}
